%% file: TVCG_2024.tex
\pdfoutput=1
\documentclass[10pt,journal,compsoc]{IEEEtran}

\usepackage{ragged2e}
\input{packages}
\begin{document}
\title{Interactive Visual Assessment \\
for Text-to-Image Generation Models}

\author{Xiaoyue~Mi,
        Fan~Tang,
        Juan~Cao,
        Qiang~Sheng,
        Ziyao~Huang,
        Peng~Li,
        Yang~Liu,
        Tong-Yee Lee
\thanks{Xiaoyue~Mi, Fan~Tang, Juan~Cao, Qiang~Sheng, and Ziyao~Huang are with the Institute of Computing Technology, Chinese Academy of Sciences, Beijing 100190,
China, e-mail: \{mixiaoyue19s,tangfan,caojuan,shengqiang18z,huangziyao19f\}@ict.ac.cn.} 
\thanks{Peng Li is with the Institute for AI Industry Research (AIR), Tsinghua University, Beijing 100084, China, e-mail: pengli09@gmail.com.}
\thanks{Yang Liu is with the Department of Computer Science and Technology, Tsinghua University, Beijing 100084, China, e-mail: liuyang2011@tsinghua.edu.cn.}
\thanks{T.-Y. Lee is with National Cheng Kung University, Tainan, Taiwan, e-mail: tonylee@mail.ncku.edu.tw.
}
}

\markboth{IEEE TRANSACTIONS ON VISUALIZATION AND COMPUTER GRAPHICS,~Vol.~XX, No.~, November~2024}%
{Mi \MakeLowercase{\textit{et al.}}: XXX}
\IEEEtitleabstractindextext{
\begin{abstract}
Visual generation models have achieved remarkable progress in computer graphics applications but still face significant challenges in real-world deployment.
Current assessment approaches for visual generation tasks typically follow an isolated three-phase framework: test input collection, model output generation, and user assessment.
These fashions suffer from fixed coverage, evolving difficulty, and data leakage risks, limiting their effectiveness in comprehensively evaluating increasingly complex generation models.
To address these limitations, we propose DyEval, an LLM-powered dynamic interactive visual assessment framework that facilitates collaborative evaluation between humans and generative models for text-to-image systems.
DyEval features an intuitive visual interface that enables users to interactively explore and analyze model behaviors, while adaptively generating hierarchical, fine-grained, and diverse textual inputs to continuously probe the capability boundaries of the models based on their feedback.
Additionally, to provide interpretable analysis for users to further improve tested models, we develop a contextual reflection module that mines failure triggers of test inputs and reflects model potential failure patterns supporting in-depth analysis using the logical reasoning ability of LLM.
Qualitative and quantitative experiments demonstrate that DyEval can effectively help users identify max up to 2.56 times generation failures than conventional methods, and uncover complex and rare failure patterns, such as issues with pronoun generation and specific cultural context generation.
Our framework provides valuable insights for improving generative models and has broad implications for advancing the reliability and capabilities of visual generation systems across various domains.
\end{abstract}
\begin{IEEEkeywords}
Visual assessment, Dynamic testing, Large language model, In context learning, Human-computer interaction
\end{IEEEkeywords}}

\maketitle

\input{Section/Introv4}
\input{Section/Related_Work}

\input{Section/Design_Motivation}
\input{Section/Methods_v4}

\input{Section/Experiments}

\section{Conclusion and Discussion}
We present DyEval, an interactive visual assessment framework that addresses the limitations of static datasets in evaluating rapidly evolving text-to-image models.
Through extensive experiments, we demonstrate that our LLM-powered approach significantly outperforms traditional evaluation methods by dynamically generating diverse test cases and providing interpretable insights into model behaviors.
The visual analytics components of DyEval enable users to effectively explore and understand complex failure patterns, particularly in challenging scenarios involving cultural nuances, implicit relations, quantifiers, and pronouns.

Our framework makes several key contributions to the visual analytics and computer graphics communities: it introduces a novel approach to adaptive model assessment that can evolve alongside advancing generation technologies; what's more, it demonstrates the effectiveness of combining LLM capabilities with interactive visualization for comprehensive model evaluation, and finally it provides a flexible foundation for analyzing various aspects of visual generation models.

\textbf{Limitations.} 
Due to resource constraints, our current evaluation focuses primarily on open-source models, and we hope to experiment with commercial models like DALL$·$E 3 to uncover more cutting-edge phenomena.
The effectiveness of the framework partially depends on the quality of the underlying LLM, which may introduce biases in test case generation. Additionally, the current implementation requires significant human involvement in the assessment process, which could be resource-intensive for large-scale evaluations.

\textbf{Future work.} 
We envision extending the framework to evaluate other visual generation tasks, particularly text-to-video generation, to support cross-modal generation systems. Another promising direction is investigating the integration of Vision-Language Models (VLMs) as a more cost-effective alternative to current LLMs for test case generation and analysis.
When text-to-image generation metrics become more advanced, we plan to incorporate them into DyEval to reduce manual evaluation effort while maintaining assessment quality.
Furthermore, we aim to develop automated bug repair mechanisms to enable self-improving generation models.

\bibliographystyle{splncs04}
\bibliography{main}
\input{Section/APPENDIX}

\end{document}

%% file: packages.tex
\usepackage{amsmath,amsfonts}
\usepackage{algorithm2e} 
\usepackage{array}
\usepackage{graphicx} 
\usepackage{pdfpages}
\usepackage{float} 
\usepackage{subcaption} 
\usepackage{textcomp}
\usepackage{stfloats}
\usepackage{url}
\usepackage{verbatim}
\usepackage{cite}
\usepackage{color}
\usepackage[T1]{fontenc} 
\usepackage{hyperref} 
\usepackage{booktabs} 
\usepackage{nicefrac} 
\usepackage{microtype} 
\usepackage{xcolor} 
\usepackage{multirow}
\usepackage{multicol}
\usepackage{times}
\usepackage{wrapfig}
\usepackage{enumitem} 
\usepackage{tcolorbox} 
\usepackage{mdframed} 
\usepackage{latexsym}
\usepackage{colortbl}
\usepackage{longtable}
\usepackage{orcidlink}
\usepackage{lipsum} 
\usepackage{xtab}

\definecolor{MocoGreen}{RGB}{57,181,74}
\newcommand{\etal}{\textit{et al.}}

\definecolor{bggreen}{rgb}{0.902, 0.984, 0.902}
\definecolor{bgblue}{rgb}{0.890, 0.996, 1.000}
\definecolor{bgpink}{rgb}{1.000, 0.886, 0.976}
\definecolor{bgyellow}{rgb}{1.000, 0.988, 0.851}
\definecolor{lightbrown}{RGB}{245, 222, 179}
\definecolor{lightgreen}{RGB}{204, 255, 204}

\RestyleAlgo{ruled} 
\LinesNumbered 

\usepackage{diagbox}
\usepackage{svg}

%% file: Section/Introv4.tex
\section{Introduction}
Recent years have witnessed remarkable advances in visual generation models~\cite{rombach2022high,podell2023sdxl,betker2023improving}.
These models show impressive capabilities in generating vivid images and have found applications across diverse computer graphics domains~\cite{10296017,10379515,10521816,10399852}.
Despite their powerful capabilities, they still exhibit various generalization failures when deployed in real-world computer graphics scenarios.
Given the vast range of images these models can generate, the potential generalization failures are highly varied and unpredictable~\cite{du2023stable}.
Thoroughly testing these models to find significant failures reliably is crucial for enhancing their performance, yet this remains a major challenge.

Current assessment approaches primarily rely on datasets and require significant manual effort, including sequential isolated three phases: the collection of test inputs, model generation, and human assessment of model outputs~\cite{saharia2022photorealistic,lee2023holistic,wang2023diffusiondb,huang2024t2i,bakr2023hrs,kirstain2023pickapic,10262331,xu2024imagereward,wu2023human}.
These fashions are widely applied but difficult to evolve.
As visual generation models advance and application scenarios expand, static datasets may become outdated, necessitating substantial resources to update and maintain their relevance.
This limited scalability also manifests in the fixed coverage of test inputs, which restricts their ability to detect model deficiencies beyond predefined scenarios.

\begin{figure} [!th]
    \centering
    \includegraphics[width=\linewidth]{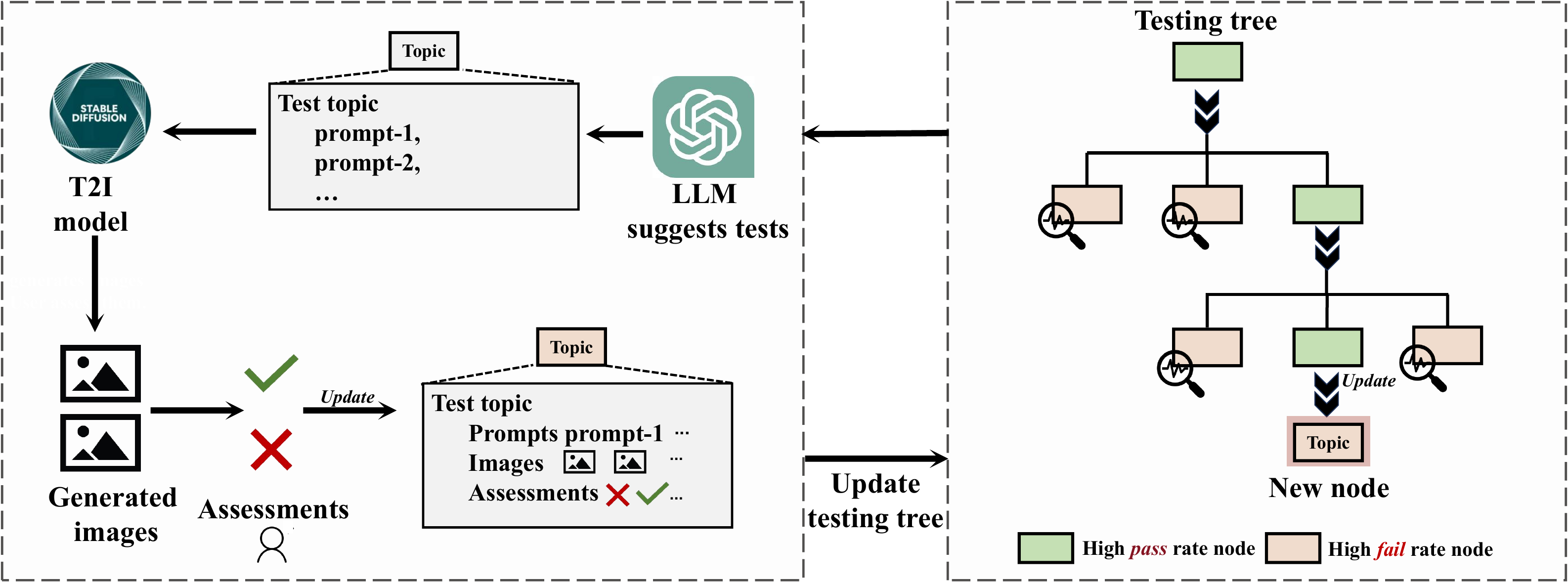}
    \caption{Diagram of the DyEval, a dynamic interactive testing framework for Text-to-Image (T2I) models. DyEval leverages a Large Language Model (LLM) to generate test prompts based on model feedback dynamically. The T2I model outputs are assessed, and results are used to update a test tree. Nodes with high pass rates (\textcolor{lightgreen}{green}) continue to explore deeper layers, while nodes with high fail rates (\textcolor{lightbrown}{brown}) are analyzed by the LLM to analyze failure reasons.
    }
    \label{fig:motivation}
    \vspace{-4mm}
\end{figure}

Interactive testing methods have emerged as a promising alternative to address above limitations.
Du~\etal~\cite{du2023stable} propose an adversarial attack method for text-to-image models,
which is adaptive based on tested model feedback but narrowly focuses on evaluating the robustness of noun changes in white-box models.
Meanwhile, open-ended human-in-the-loop testing~\cite{ribeiro-etal-2020-beyond,ganguli2022red,ribeiro2022adaptive,Gao_2023_ICCV} leverages Large Language Models (LLMs) and interactions with evaluators to generate challenging data for testing models on coherent, manually specified topics.
Although effective in text-input text-output tasks or image understanding tasks, these methods cannot be directly applied to the visual generation domain due to the complexity of visual generation failures~\cite{borji2023qualitative}.
Evaluators will have a huge cost to mark failures of generated images in detail.

In this paper, we introduce DyEval, a novel interactive visual assessment framework that leverages an LLM through in-context learning to adaptively generate test inputs based on model feedback, effectively identifying failures in open domains (Fig.~\ref{fig:motivation}).
DyEval features an intuitive visual interface that enables users to interactively explore and analyze model behaviors while maintaining a comprehensive overview of the testing process.
DyEval is built upon a tree-based structure systematically recording the testing process. Each tree node contains test topics, specific prompts as test inputs, and corresponding evaluation results. The framework operates through two main components:
First, in the test node construction phase, the LLM generates test inputs based on the current test topic and existing contexts. Evaluators then assess the generated images through the interactive interface, determining whether they pass or fail the specified criteria. 
Second, DyEval performs adaptive exploration based on the evaluation results. For nodes with high pass rates, the LLM proposes new test topics for deeper tree exploration. For nodes with lots of failures, DyEval employs contextual reflection, combining dynamic failure location to identify failure-inducing parts with a self-reflection module analyzing potential failure types and reasons based on existing test contexts.
Benefiting from the dynamic interactivity of DyEval, we can also avoid the risk of data leakage~\cite{magar2022data,oren2024proving}, which typically arises from public static test sets leading to targeted model optimizations and, consequently, skewed evaluation results.

Experimentally, we showcase the efficacy of DyEval by evaluating various state-of-the-art text-to-image models from four perspectives: Object, Relation, Attribute, and Context (global attributes in the image, such as style).
Quantitative experiments demonstrate that DyEval significantly outperforms traditional static evaluation methods, identifying max up to 2.56 times more failure cases when evaluating the same number of text-image pairs.
Our experiments also reveal consistent patterns across different generation models.
All tested models perform better with material objects over abstract ones, static over dynamic attributes, and explicit over implicit relations. While excelling at style-related tasks, they struggle with culture and knowledge-based generations, particularly with cultural nuances and implicit relations.
In Parts-of-Speech~(PoS) analysis of text inputs, we find SDXL and SD3 show improvements over SD1-5 and SD2-1, but they continue to struggle with specific linguistic elements such as quantifiers and pronouns.
Through case studies, DyEval uncovers unexpected testing perspectives and intricate failure patterns.
The contextual reflection module identifies specific triggers for model failures, such as culturally specific words (e.g., ``kimono'') and certain text combinations. These findings provide valuable insights into the current limitations and potential areas for improvement in text-to-image generation models.
In summary, DyEval offers a novel perspective in text-to-image model testing, significantly aiding evaluators in comprehensively understanding the boundaries of model capabilities and inspiring future enhancements.
In this paper, our main contributions
are listed below:
\begin{itemize}
    \item We propose DyEval, a novel dynamic interactive visual assessment framework for adaptive testing of text-to-image models, that enable efficient human-in-the-loop assessment and analysis.
    DyEval can flexibly adapt to any user-defined criteria like alignment, bias, and more, for both white-box and black-box models.
    \item With the LLM-powered iteration system and interactive visual interface, DyEval gradually adapts tests based on user feedback to find a variety of model failures.
    Compared with static testing methods~\cite{saharia2022photorealistic，huang2024t2i,10262331}, DyEval allows limitless test scope, granularity expansion, and coevolution with evolving tested models.
    \item Experiments validate that DyEval effectively finds the complex and uncommon failures of tested models, such as pronoun-induced degradation, which provides valuable and interpretable insights for further improvement.
\end{itemize}

%% file: Section/Related_Work.tex
\section{Related Works}
\subsection{Assessments of Visual Generation Models}
In the field of visual generation, there has been a significant increase in evaluation studies over the past years, especially in text-to-image generation.
Currently, static evaluation~\cite{saharia2022photorealistic,lee2023holistic,wang2023diffusiondb,huang2024t2i,bakr2023hrs,kirstain2023pickapic,10262331,xu2024imagereward,wu2023human,10693493} are the predominant method for evaluation, which employs a one-time assessment using pre-collected data sets, which can show the generalization ability of the model on specific data.
Firstly, evaluators determine testing perspectives such as concept conjunction~\cite{liu2022compositional}, and spatial relationships~\cite{wu2023harnessing}.
Then, they gather different test prompts according to their testing perspectives.
There are four main ways that models get their prompts: existing text-image datasets~\cite{lin2014microsoft}, human usage records from online text-to-image tools like Midjourney~\cite{saharia2022photorealistic,wang2023diffusiondb,kirstain2023pickapic}, semi-automated generation using LLMs with predefined templates~\cite{bakr2023hrs,huang2024t2i}, and combinations of them~\cite{lee2023holistic,xu2024imagereward,wu2023human,10262331}.
Finally, evaluators assess the model outputs using established metrics and manual evaluation.
Due to the high challenge of image generation tasks, manual assessment is indispensable and remains the most reliable~\cite{jiang2024genai}.

These works only provide static evaluations, they are unable to adapt to evolving models and human needs.
Du~\etal~\cite{du2023stable} propose a gradient-based adversarial attack method for text-to-image models, dynamically investigating the adversarial robustness of model inputs by adding or changing nouns.
However, their approach is hard to generalize in the open domain.

In this paper, we offer a novel adaptive visual assessment method that enables users to specify test themes~(e.g. culture, knowledge, count, and more), test aspects~(e.g. alignment, bias, and more), and test metrics~(e.g. Clip score, human assessment, and more) in the open domain and adjust the testing process interactively based on the model assessment feedback.
This evaluation approach is useful in accommodating the open domain and evolving capabilities of the text-to-image models and can be seen as complementary aspects of current evaluations in text-to-image models, jointly enhancing the reliability and practical utility of the models.

Several studies~\cite{lu2023llmscore,chen2023x,hu2023tifa,cho2024davidsonian} have utilized large language models~(LLMs) as metrics to evaluate text-image alignment in text-to-image tasks. For instance, LLMScore~\cite{lu2023llmscore} converts images into both image-level and object-level visual descriptions, subsequently employing LLMs to assess the alignment between the generated images and their corresponding texts through evaluation instructions.
X-IQE~\cite{chen2023x} uses large visual language models to generate textual explanations of generated image and text alignment.
Several studies~\cite{hu2023tifa,cho2024davidsonian} employ Visual Question Answering (VQA) to assess the alignment between text and images. In these approaches, a language model is utilized to automatically generate multiple question-answer pairs based on a given text input. The faithfulness of the image is then evaluated by determining whether existing VQA models can accurately respond to these questions using the corresponding image.

Different from focusing on text-image alignment metrics, DyEval is an interactive assessment protocol, and decoupled from specific aspects and assessment metrics.
\subsection{Human-LLM Collaborative Model Testing}
Leveraging LLMs for human-aided evaluation, i.e. open-ended human-in-the-loop testing, has been well-recognized in text-input text-output tasks~\cite{ribeiro-etal-2020-beyond,ganguli2022red,ribeiro2022adaptive} and image-input understanding tasks~\cite{Gao_2023_ICCV}.
They adopt strategies from software engineering, engaging individuals to generate test scenarios with assistance from LLMs.
Utilizing human intervention to explore input scenarios beyond conventional training and validation datasets, this testing methodology has effectively identified consistent failures in state-of-the-art models, even performing unproblematic on static benchmarks.

For text-input text-output tasks, they manually specify testing topics, using the powerful text generation, rich knowledge, and logical reasoning abilities of LLMs to create test inputs.
Then, evaluators select and evaluate model errors manually and use this feedback to generate new test inputs.
Similarly, text-to-image is also a text-input task, and its evaluation heavily depends on manual assessment, with no similar testing methods yet in the field.

However, these methods cannot be directly applied to the text-to-image domain due to the complexity of image generation failures.
Considering the complex dimensions outputs result in many types of failure~\cite{borji2023qualitative} in image generation tasks, it would be highly costly for humans to detailly mark these failures.
In this paper, we explore human-aided evaluation in the text-to-image task and introduce a contextual reflection module to alleviate this problem by automatically identifying minimal failure test inputs and conducting analysis based on LLMs.

%% file: Section/Design_Motivation.tex
\section{DESIGN OBJECTIVES}
Based on our analysis of limitations in current visual assessment fashions, we identify four key design objectives for our visual assessment framework for text-to-image models:

\textbf{O1 Dynamic coverage.} Our visual assessment framework should overcome the fixed coverage limitation of static testing by supporting open-ended exploration of test scenarios.
The ideal framework should enable visual exploration of dynamically generated test cases across different aspects.
Through human assessment feedback, Our framework can interactively adjust testing granularity based on discovered failure patterns.
The visualization framework should support visual navigation between breadth-first exploration for comprehensive coverage and depth-first investigation for specific issues, enabling users to continuously expand the testing boundary without being constrained by predefined scenarios.

\textbf{O2 Evolutionary adaptability.} To address the evolution difficulty in static datasets, the visual assessment framework should facilitate dynamic visualization of testing strategies that adapt based on model feedback and user needs.
Through interactive visualizations and visual feedback loops, users can continuously refine test cases without requiring complete dataset reconstruction, ensuring the testing process remains effective as models advance.

\textbf{O3 Anti-leakage design.} To prevent the data leakage issues common in static testing, the visual assessment framework should generate test inputs dynamically rather than fixed test sets.
And test input distribution is broad enough to make it difficult to target optimization overfitting even if one knows how the test inputs are generated.
Through visual monitoring of customizable testing criteria that evolve with model capabilities, the framework maintains testing effectiveness while preventing models from being specifically optimized for known test cases.

\textbf{O4 Visual scalability.} The framework should facilitate efficient visual analysis of large-scale testing results through interactive visualization of hierarchical test structures at multiple levels of detail.
The framework should support visual exploration and pattern discovery of failure triggers through coordinated views, while providing clear LLM-generated insights and failure analysis.
The interface should maintain overview and detail-on-demand capabilities while ensuring framework responsiveness and clarity.

%% file: Section/Methods_v4.tex
\begin{figure*}[htbp]
    \centering
    \includegraphics[width=1.0\linewidth]{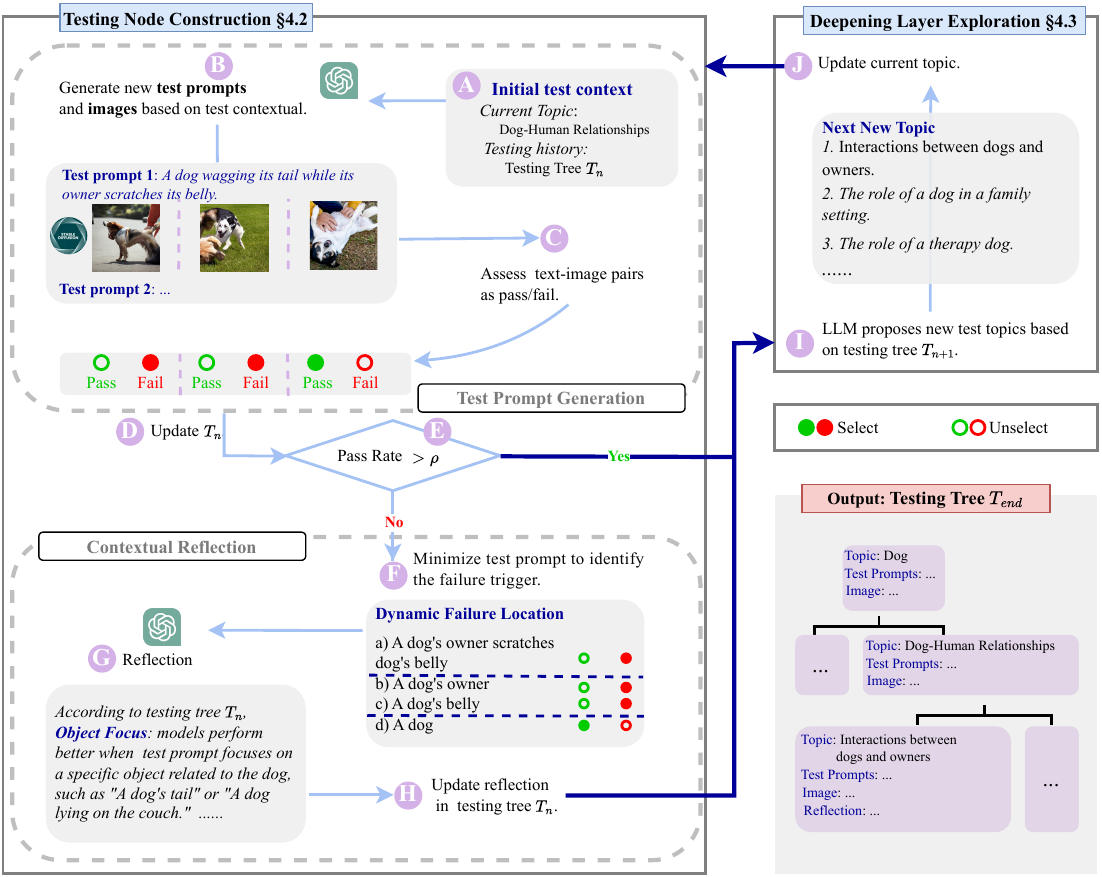}
    \caption{Overview of DyEval.
    The DyEval process begins with setting initial topics and model parameters (Step A). In the test input generation phase, the LLM creates initial test inputs for the selected topic (Step B). Evaluators then review and annotate the generated images (Step C) and update the test tree accordingly (Step D). If a test node shows low average pass rates (Step E), the LLM refines these inputs to identify potential failure triggers in the dynamic failure location module (Step F). It then reflects on all available information related to the current topic (Step G) and updates the test tree (Step H). If the pass rates are satisfactory, the LLM proposes new topics based on the test context (history testing records in the testing tree) (Step I). Evaluators can select these new topics (Step J) to generate additional test inputs, continuing the evaluation cycle. This iterative process continues until the maximum exploration depth is reached. 
    }
    \label{fig:pipeline}
\end{figure*}
\section{Methods}
In this section, we present the formulation and overall pipeline in Sec.~\ref{sec:for}. Subsequently, we provide detailed descriptions of the two primary steps of DyEval: testing node construction (Sec.~\ref{Sec:input}) and deepening node exploration (Sec.~\ref{Sec:topic}).

\subsection{Overview}
\label{sec:for}
\textbf{Formulation.}
A $test$ is defined as a combination of a text input $i$ and the expected outputs $X$ of a text-to-image model $m$ that meet evaluator preferences and intentions~\cite{ribeiro-etal-2020-beyond, ribeiro2022adaptive, Gao_2023_ICCV}.
These preferences include aesthetic appeal, semantic alignment, level of detail, fairness, and more.
We define a $bug$ as a specific text input where the visual generation model consistently fails to generate an accurate corresponding image.
For each test pair $(x, i)$ where $x \in X$, the $\text{test}(x, i)$ passes if $x$ meets the evaluation expectations; otherwise, it fails. Assuming a distribution of output images given test inputs $\mathbf{P}(X|i,m)$, a test$(i)$ fails when the pass rate is lower than the evaluator-expected pass rate $\rho$, termed as a $bug$:
\begin{equation}
    \mathbb{E}_{x \sim \mathbf{P}(X \mid i,m)}[\operatorname{test}(x, i) \text{ passes }] < \rho.
\end{equation}

We then define a $topic$ as a coherent group of tests whose text prompts are united by an explainable concept~\cite{eyuboglu2022domino,wiles2022discovering} and share similar expectations~\cite{ribeiro2022adaptive, Gao_2023_ICCV}. Each topic$(t_o)$ corresponds to a set of text inputs $I$ that fit the specific topic $t_o$ (e.g., "dogs"). For example, $I$ contains text inputs related to dogs, and $i \in I$. The text-to-image model $m$ generates output images $X$ based on these inputs:
\begin{equation}
    \text{topic}(t_o) = \{\text{test}(x,i \mid t_o, m) \mid i \in I, x \in X \}.
\end{equation}
DyEval aims to assist evaluators in revealing test topics and test inputs with high failure rates, thereby accurately finding the capability boundaries of the tested model.

The full process of DyEval is recorded as a test tree $T$ with a maximum test depth $d_{\text{max}}$ and a maximum test width $w_{\text{max}}$, which also represents test contexts. A test node $t_{d,w} \in T$ is defined as:
\begin{equation}
    t_{d,w} = \{ t_{o_{d,w}},\text{topic}(t_{o_{d,w}}), \text{result}_{d,w}, r_{d,w} \},
\end{equation}
where $d$ represents the index of depth, $w$ represents the index of width, $r$ signifies the reflection conducted by the LLM for each node, and $\text{result}_{d,w}$ denotes the assessment results of $\text{test} \in \text{topic}_{d,w}$. For convenience, we also define $f_{d,w}$ as the parent node of $t_{d,w}$ and $ch_{d,w}$ as the children nodes of $t_{d,w}$.
Additionally, several hyperparameters must be set for testing, including the number of test topics generated per iteration $n_t$, the number of test inputs generated per iteration $n_i$, and the number of images generated per test input $n_x$.

\textbf{Pipeline.}
As illustrated in Fig.~\ref{fig:pipeline}, DyEval implements our design objectives through a visually explorable test tree structure:
To support \textit{dynamic coverage} (O1), evaluators begin by setting an initial test topic $topic_{0,0}$ (e.g., ``DOG'' in Fig~\ref{fig:pipeline}) that serves as the root node for subsequent exploration. The visual interface allows users to flexibly navigate and expand the testing space from this starting point.

Following the \textit{evolutionary adaptability} principle (O2), DyEval constructs testing nodes through an interactive process (Sec.~\ref{Sec:input}). The LLM dynamically generates text inputs $I$ based on the current node topic and accumulated test contexts. The text-to-image model $m$ produces images $X$ for testing, which are visually presented for evaluator assessment against predefined criteria.
For nodes with low average pass rates (below threshold $\rho$), DyEval transitions into a contextual reflection module.
This module helps identify failure triggers, i.e. the minimal input subsets leading to failures, and enables the LLM to analyze possible failure patterns based on related test contexts.
All these processes are visually logged and accessible through the interactive interface.

To ensure \textit{anti-leakage design} (O3) and support \textit{dynamic coverage} (O1), DyEval continuously generates new test scenarios rather than relying on fixed test cases.
For nodes with high average pass rates, DyEval supports deeper layer exploration (Sec.~\ref{Sec:topic}). The LLM generates new test topics dynamically, and evaluators can interactively select topics to pursue, creating unique testing paths that prevent model optimization for specific test cases.
This iterative loop continues until the maximum predefined testing depth is reached.

Supporting \textit{visual scalability} (O4), the entire process is organized in a hierarchical test tree structure that allows efficient navigation and analysis at multiple granularities. The visual interface maintains both overview and detailed views of the testing progress, enabling evaluators to track the exploration process until reaching the maximum predefined testing depth while managing complexity through interactive visualizations.
\subsection{Testing Node Construction}
\label{Sec:input}
\subsubsection{Testing Prompt Generation}

In constructing a testing node, DyEval begins by utilizing the strong text generation ability of LLM to generate $n_i$ text prompts $I$ based on the current topic.
As illustrated in Fig.~\ref{fig:pipeline}, these prompts correspond to the current test topic, such as ``DOG-human relationships.'' An example of a generated prompt is ``A dog wagging its tail while its owner scratches its belly.''
If the test tree $T$ includes additional information related to the current topic, such as parent nodes $f$, these are also incorporated as test contexts for the LLM.
DyEval inspires the LLM to generate test inputs that are semantically aligned with the topic, aiming to identify new potential failures in the model being tested.
This can be formulated as:
\begin{equation}
    I_{d,w} = \text{InputGen}(t_{o_{d,w}}, f_{d,w}),
\end{equation}
where $InputGen$ is a process that LLM takes the current topic $t_{o_{d,w}}$ and parent node $f_{d,w}$ as test contexts input and output testing inputs for text-to-image models. 
The prompt details for this function are provided in the supplementary material.
To maintain semantic relevance, we verify the generated test inputs with the LLM and eliminate duplicates using an n-gram matching algorithm.

After the generation of test inputs, the tested model $m$ generates $n_x$ images per input, and evaluators will determine \textcolor{green}{pass} or \textcolor{red}{fail} based on the specific criteria.
Given the complexity of image generation tasks, manual evaluation is essential and remains the most reliable method~\cite{jiang2024genai}.
To reduce the labeling effort, we use CLIPScore~\cite{hessel2021clipscore} as an initial filter; any text-image pair with a CLIPScore below a certain threshold is automatically marked as a fail.
To avoid mislabelling these results are also shown to the users, and they can refine as needed.
Finally, these test processes are systematically recorded in the test tree $T$, with updates made to the relevant nodes $t_{d,w}$ within $T$.
\subsubsection{Contextual Reflection}
\label{Sec:reflection}

For low average pass rate test nodes, we design a contextual reflection module to analyze further possible failure patterns of the text-to-image model $m$.
This module consists of two components: a dynamic failure location that attributes failures to the smallest text component, called a failure trigger, and self-reflection for summarization and analysis by the logical ability of LLM. 

\textbf{Dynamic failure location.}
When a test $\text{test}(x,i)$ fails, it is both costly and highly subjective for evaluators to label failures in detail.
However, analyzing failure reasons and exploring the weaknesses of the test model is crucial for understanding and improving the model.
To address this, we devise a divide-and-conquer strategy, called dynamic failure location, as illustrated in Fig.~\ref{fig:pipeline} and Alg.~\ref{algorithm:mini}. 
This approach iteratively breaks down the text input $i$ to pinpoint failure triggers.
Initially, the input is split into two halves and tested. If both halves fail, further splitting continues. If both pass, one half is selected for further splitting and merged with the other one separately for testing.

To facilitate precise analysis and control of the text input and ensure semantic coherence after decomposition, we employ an LLM to transform text input $i$ into a scene graph representation $c$~\cite{johnson2015image}.
Scene graph, as a structured representation method, explicitly describes objects, their attributes, and relationships between objects within a scene.
Converting visual content into scene graphs represents a crucial goal in visual understanding.
The process $F_{test}$ subsequently transforms these scene graphs back into text inputs for evaluating text-to-image models, which is implemented using LLM..

As detailed in Alg.~\ref{algorithm:mini}, after converting the test input $i$ into scene graph $c_0$, dynamic failure location divides it into two subsets, $c_1$ and $c_2$. These subsets are then tested by converted into texts to determine if they meet evaluator requirements.
If a subset fails, it undergoes further subdivision. If both pass, one is retained as a baseline while the other is further subdivided. All results are stored in the test node $t$ for future reflection. Additional examples and specific task prompts of $F_{test}$ and $F_{split}$, and other details are available in the supplementary material.

\input{Algorithm/mini-input}

\textbf{Self-reflection.}
Leveraging the logical reasoning and analysis ability of LLM~\cite{yao2022react, shinn2023reflexion}, DyEval tries to analyze and summarize failure patterns and reasons for tested model $m$.
The reflection, denoted as $r_{d,w}$, is generated by the LLM using the current test node $t_{d,w}$ from the test tree $T$:
\begin{equation}
    r_{d,w} = \text{Reflect}(t_{d,w}),
\end{equation}
where $Reflect$ is a process where the LLM uses all available information from the current node $t$ to analyze potential failure patterns.
This reflective process allows DyEval to conduct a thorough analysis of failed test data, offering a deeper understanding of model performance and identifying potential areas for improvement.
The specific prompt used is detailed in the supplementary materials.

\subsection{Deepening Layer Exploration}
\label{Sec:topic}

For test nodes with a high average pass rate, the next step involves adaptively generating further exploration topics based on the current test context.
While labeling text-image pairs is easy for humans, generating new topics is challenging and cost high.
Therefore, we offload this creative task to an LLM, as illustrated in Fig.~\ref{fig:pipeline} on the right.
The capabilities of LLM in language generation, knowledge, and summary analysis are leveraged to generate suggested test topics for the next layer using the current testing node $t_{d,w}$ as context:
\begin{equation}
    t_{o_{d+1}} = \text{TopicGen}(t_{d,w}),
\end{equation}
where $t_{o_{d+1}}$ is a set of $n_t$ test topics for children nodes of $t_{d,w}$.
These new topics automatically are created as new nodes in $T$ with parent node $t_{d,w}$ and updated as the new nodes $t_{d+1}$.
By default, DyEval recommends test topics based on breadth-first, but evaluators can manually change the order for further exploration in the next testing loop.
For instance in Fig.~\ref{fig:pipeline}, after testing the node on the topic ``DOG-human relationships'', the LLM suggests a new set of child topics, such as `` Interactions between dogs and owners.'', ``The role of a dog in a family setting.'', and ``The role of a therapy dog.''.

By deepening layer exploration, DyEval facilitates the generation of subsequent test topics and inputs, guided by the evolving test tree $T$, until reaches a predefined limit of testing loops.
This approach broadens the testing scope by suggesting more fine-grained topics related to the current one, enabling evaluators to refine their focus for future tests.

%% file: Algorithm/mini-input.tex
\begin{algorithm}[htb]
\caption{
Dynamic Failure Location.
}
\label{algorithm:mini}
\KwIn{
Initial scene graph $c_0$,
Split function $F_{split}$,
Test function $F_{test}$,

}
\KwOut{
Process test results $T_{mini}$.
}
$T_{mini} \leftarrow \{(c_0, fail)\}$ \;
$t_{mini} \leftarrow \emptyset $ \;
$Q \leftarrow \{(c_0, \emptyset)\}$ \;
\While{ $Q \neq \emptyset $}{
    $c, l \leftarrow Q.pop() $ \;
    \uIf {$|c| \neq 1$} {
        $c_1, c_2 \leftarrow F_{split}(c)$ \;
        $test(c_1 \cup l) \leftarrow F_{test}(c_1 \cup l)$ \;
        $test(c_2 \cup l) \leftarrow F_{test}(c_2 \cup l)$ \;
        $T_{mini}.append((c_1 \cup l, test(c_1 \cup l)))$\;
        $T_{mini}.append((c_2 \cup l, test(c_2 \cup l)))$\;
        \uIf {$test(c_1 \cup l) \; fails$} {
            $Q \leftarrow Q \cup \{(c_1, l)\}$ \;
        }
        \uIf {$test(c_2 \cup l) \; fails$} {
            $Q \leftarrow Q \cup \{(c_2, l)\}$ \;
        }
        \uIf { $test(c_1 \cup l) \; passes \wedge test(c_2 \cup l) \; passes$ } {
            $Q \leftarrow Q \cup \{(c_1, c_2\cup l), (c_2, c_1\cup l)\}$ \;
        }
        $t_{mini} \leftarrow t_{mini} \cup \{ test(c_1 \cup l), test(c_2 \cup l) \} $ \;
    }
}
return $T_{mini}$\;

\end{algorithm}




%% file: Section/Experiments.tex
\section{Experiments}
In this section, we start by introducing experimental settings in Sec.~\ref{sec:ex_setup}, and then conduct both qualitative and quantitative experiments on DyEval to answer the following questions:
(1) Can DyEval find bugs effectively in the models being tested, especially compared with static methods (Sec.~\ref{sec:ex_useful})?
(2) What are the differences and similarities in failures among various text-to-image models (Sec.~\ref{sec:ex_difference})?

\subsection{Experimental Settings}
\label{sec:ex_setup}
\textbf{Hyper-parameters of DyEval.}
We evaluated four major open-source text-to-image models: Stable Diffusion v1-5 (SD1-5)~\cite{rombach2022high}, Stable Diffusion v2-1 (SD2-1)~\cite{rombach2022high}, Stable Diffusion XL (SDXL)~\cite{podell2023sdxl}, and Stable Diffusion v3 (SD3)~\cite{esser2024scaling}.
Our study involved 23 participants, all proficient in English, without color blindness, and trained for the task.
And remuneration to participants in excess of the local minimum hourly wage.
\textcolor{black}{In our participant evaluation process, we randomly reintroduce previous text-image pairs to test annotation consistency.
The inter-rater reliability analysis of participants' assessments on text-image pairs yielded Kendall's tau coefficient of $\tau = 0.7746$ ($p < 0.05$ ), indicating strong agreement among raters after training.}
The parameters used were: number of test topics per iteration $n_t = 3$, number of test inputs per iteration $n_i = 5$, number of images generated per test input $n_x = 4$, and max depth $d_{max} = 3$.
We set the topic stop extension pass rate $\rho$ to 0 to analyze text-to-image models fully.
The LLM we used is GPT-3.5-turbo which is developed by OpenAI and used widely in agent and LLM applications~\cite{yao2022react,shinn2023reflexion}.
Detailed prompts in the supplementary material.

\textbf{Assessment criteria.}
Here we set the assessment criteria as text-image alignment and visual quality.
A $test(x,i)$ passes if the generated image aligns with the text input $i$ without obvious defects, missing elements, misplacements, or extraneous content that was incongruent with the text input.
Conversely, a $test(x,i)$ fails if these criteria are not met.
Note that, DyEval is a dynamic interactive testing protocol that can be integrated with any evaluation criteria and assessment methods.

\textbf{Evaluation metrics.}

We define the number of bugs found as $\#Bugs$, the average pass rate as APR, and the average fail rate as $AFR$. These can be formalized as follows:
\begin{equation}
\text{\#Bugs} = \sum_{t\in T} \#\text{test}(i)~\text{fails}, \text{test} \in t,
\end{equation}
\begin{equation}
\text{APR} = \frac{\#\text{test}(x,i)~\text{passes}}{\#\text{test}_{total}},
\end{equation}
\begin{equation}
\text{AFR} = 1 - \text{APR},
\end{equation}
where ``\#'' is an abbreviation for ``the number of'', and $\#\text{test}_{total}$ is the total number of test pairs in a testing tree $T$ (which do not contain contextual reflection modules).
By analyzing both APR/AFR and \#Bugs together, we can gain a more nuanced understanding of model performance.
The average rate APR/AFR provides a general performance overview, while the number of poorly performing cases (\#Bugs) highlights specific weaknesses.
This dual approach allows for a balanced evaluation, ensuring that models are not only generally effective but also robust against challenging scenarios.

\textbf{Initial topic settings.}
Building on prior visual assessment frameworks~\cite{lee2023holistic,huang2024t2i,bakr2023hrs,kirstain2023pickapic,xu2024imagereward} and image scene graph works~\cite{johnson2015image}, we encompass a comprehensive assessment from four perspectives: Object, Relation, Attribute, and the global attribute Context, which is crucial for understanding the broader implications of image generation. Here Context refers to the broader setting or environment that influences how objects, relations, and attributes are perceived. It could include cultural, historical, or situational factors that provide additional meaning to the image. Each perspective is further detailed as follows:

\begin{itemize}
    \item \textbf{Object}
    \begin{itemize}
        \item \textit{Material Objects}: These are tangible entities that exist in the physical world, such as animals, vehicles, and furniture.
        \item \textit{Abstract Objects}: These refer to conceptual entities that do not have a physical form, such as emotions, ideas, and philosophies.
    \end{itemize}
    
    \item \textbf{Relation}
    \begin{itemize}
        \item \textit{Explicit Relations}: These are directly observable relationships within images, such as spatial arrangements and object interactions.
        \item \textit{Implicit Relations}: These require contextual understanding and knowledge, such as social relationships or implied connections.
    \end{itemize}
    
    \item \textbf{Attribute}
    \begin{itemize}
        \item \textit{Static Attributes}: These are inherent characteristics of objects that remain constant, such as shape, color, and texture.
        \item \textit{Dynamic Attributes}: These describe characteristics that can change over time, such as movement, growth, and transformation.
    \end{itemize}
    
    \item \textbf{Context}
    \begin{itemize}
        \item \textit{Style}: This refers to the overall artistic or visual style that influences the perception of the image, such as impressionism or realism.
        \item \textit{Culture}: This involves cultural elements that provide background context, such as traditional symbols or practices.
        \item \textit{Knowledge}: This encompasses the broader understanding of historical or situational contexts that inform the image's interpretation.
    \end{itemize}
\end{itemize}

These categories allow for a nuanced evaluation of text-to-image models, highlighting areas where current benchmarks may fall short, particularly in handling abstract concepts, implicit relationships, and cultural nuances.

\subsection{Usefulness of DyEval}
\label{sec:ex_useful}
In this experiment, we explored nine initial topics: material objects, abstract objects, explicit relations, implicit relations, static attributes, dynamic attributes, style, culture, and knowledge.
Participants used DyEval to generate three subtopics for each topic, continuing until they reached a maximum depth of three, and we could collect a maximum of 13 test topics per test tree (1 test node at depth 1, 1$\times$3 test nodes at depth 2, and 3$\times$3 test nodes at depth 3).
We then cleaned the data to delete user errors, such as duplicates or omissions, and conducted our analysis on this cleaned data.

\input{Figures/Experiment1}

\textbf{Accumulated number of bugs.}
Fig.~\ref{fig:exp2} shows the average accumulated number of bugs found by DyEval during testing, which reveals that DyEval can consistently find bugs in the models.
As testing progresses, the variance in  $\#$Bugs across different topics increases, as shown by a growing standard deviation.
When comparing the final  \#Bugs and the speed of bug finding, the ranking is clear: SD1-5> SD2-1 > SD3, SDXL.
It indicates that SD3 and SDXL have higher generation performance than SD1-5 and SD2-1, and are less prone to trigger bugs.
Fig.~\ref{fig:ablation} and Fig.~\ref{fig:exp1} further confirm that model performance follows the trend SD3, SDXL > SD2-1 > SD1-5, aligning with current expectations in the field.

\textbf{Finding 1: Models with superior performance are more challenging to find their bugs.}
This finding underscores the reliability and usefulness of DyEval, highlighting the inherent challenge of identifying bugs in high-performing models.
This insight is crucial for developers aiming to deploy robust text-to-image models in real-world applications, where undetected bugs could lead to significant issues.
\textcolor{black}{Notably, while SD3 exhibits a higher cumulative \#Bugs, its overall pass rate is better according to Fig.~\ref{fig:exp1}.
This indicates that SD3 maintains more consistent performance across different test inputs, whereas SDXL shows more extreme variations in its results—achieving higher success rates on favorable cases but performing poorly on challenging ones.
Here, bugs are defined as test inputs with an average pass rate below 0.75, and SD3 has many inputs with pass rates between 0.5 and 0.75.}

\input{Figures/Ablation1}

\textbf{Comparation with non-adaptive testing.}
We evaluate the efficacy of the DyEval framework against a non-adaptive testing framework, where the LLM generates the same quantity of text inputs directly~(65 be exact, 13 nodes$\times$ 5 test inputs per node, excluding prompts in the contextual reflection module for our method).
This evaluation was conducted across four distinct models to assess the failure-finding capabilities of each method.
We calculated the average pass rate APR for each model across four relatively objective initial topics: ``material objects'', ``explicit relations'', ``static attributes'' and ``style''.
We avoid selecting ``culture'', ``implicit relationships'', or other subjective initial topics due to their complexity and lower human assessment agreement. 

As shown in Fig.~\ref{fig:ablation}, our experimental results show that when assessing an equal number of text-image pairs, DyEval detects more bugs compared to conventional static evaluation approaches (max up to 2.56 times in SD1-5).

Moreover, DyEval finds more differentiated test cases, as the APR difference among SD1-5, SD2-1, SDXL, and SD3 is larger under the DyEval test.
This indicates a superior testing efficiency of the DyEval framework in finding model vulnerabilities across different text-to-image models and testing aspects.
Finally, DyEval shows a smaller variance across test themes for more stable testing than static testing.

\textbf{Case study.}
\input{Figures/Case}
\input{Figures/ladar}

As illustrated in Fig.~\ref{fig:case}, we present a complete testing tree for SD3, starting with the initial topic ``Spatial relationships''.
Each node in the tree indicates its specific test topic and the corresponding average pass rate, demonstrating that DyEval can progressively explore the boundaries of SD3.
Beginning with ``Spatial relationships'', DyEval delves into finer-grained testing scenarios such as ``Object orientation'', ``Proximity and distance'', and ``Size comparison and scaling''.
This detailed exploration helps identify the model's capabilities and limitations across various test topics.
During this test process, we can discover that SD3 encounters challenges in scenarios like ``Object orientation in complex environments'', ``Depicting emotional distance through body language'', and ``Relative size comparison with animals''. Interestingly, the average pass rate of child nodes is not always lower than that of their parent nodes.
This occurs because child nodes represent more specific scenarios within the broader topic of the parent node, which the model might handle better. Thus, exploring the model's boundaries involves recognizing both its weaknesses and strengths.

Additionally, Fig.~\ref{fig:case} showcases specific test inputs along a testing path.
It reveals that when generating child test nodes, DyEval considers the test records of parent nodes.
For instance, failures in generating a cat and mouse led to the child topic ``Size comparison and scaling'', and the successes in generating mountains and villages, contrasted with failures in generating whales and ships, resulted in a new child test topic ``Relative size comparison with animals''.

\input{Figures/Experiment2}

\input{Figures/Error_analysis}
\input{Figures/error_analysis_Case}
\subsection{Similarity and Difference Among Different Models}
\label{sec:ex_difference}

\textbf{APR of different models across topics.}
Fig.~\ref{fig:exp1} shows the average pass rate $APR$ of SD1-5, SD2-1, SDXL, and SD3 across nine initial topics.
Our experiments reveal consistent patterns in generation performance across different models, with similar relative strengths and weaknesses in specific areas.

In object generation, models demonstrate stronger capabilities with abstract objects compared to material ones.
This is because abstract objects rely on conceptual and emotive elements, which are less restricted by physical characteristics and evaluators often assess them based on subjective interpretation and emotional resonance.
In contrast, material objects are evaluated against more stringent criteria, emphasizing detail and authenticity grounded in real-world experiences.
This disparity in evaluation standards highlights the inherent challenges models face when tasked with generating material objects.

Regarding attributes, models excel at generating static characteristics such as color, shape, and texture, benefiting from abundant training data and their unchanging nature.
However, they struggle with dynamic attributes involving temporal changes or movement, revealing limitations in representing time-dependent visual characteristics within single-frame generations.

Furthermore, the models exhibit varying degrees of proficiency in relation generation.
They struggle with implicit relations, which points to gaps in their textual comprehension and knowledge integration, particularly in scenarios involving human interactions.
Explicit relations are generally handled more effectively, suggesting that models benefit from clear and direct contextual cues.

When it comes to context generation, models are adept at capturing style through visual elements such as lines, colors, and shapes. Knowledge-based content, which necessitates background information, is enhanced by explicit labels and descriptions present in the training data.
Nonetheless, cultural content remains a significant challenge due to its complexity, requiring a nuanced understanding of symbols, customs, and traditions across diverse cultural contexts.

\textbf{Finding 2: Different versions of text-to-image models share consistent performance patterns, with notable weaknesses in culture and implicit relationship handling.}
These observations underscore the capabilities and limitations of the models, highlighting the need for further research to enhance their understanding and generative performance, particularly in areas rich in cultural nuances and implicit relational contexts.

\textbf{Part-of-Speech (PoS) Analysis.}
We analyze test input data from Sec.\ref{sec:ex_useful} excluding data generated by the contextual reflection module, group them by tested model, and divide them into fail and pass sets.
After removing stop words and infrequent PoS categories, we calculate the average failure rate AFR for each PoS across models.
Fig.\ref{fig:pos_tagging_analysiy} displays the PoS rankings by AFR. 
Please refer to the supplementary material for detailed PoS meanings and distribution across models.

For performance comparison, SD3 and SDXL generally outperform SD2-1 and SD1-5 across various PoS categories. The AFR ranking is SD1-5 > SD2-1 > SDXL > SD3 from highest to lowest. SDXL and SD3 show significant improvement in clause and verb comprehension compared to earlier models, as evidenced by lower AFR in the WRB (Wh-adverb) and VB (Verb) series.
SD3 and SDXL exhibit different strengths.
SD3 shows lower failure rates in CD, RB, WDT, and CC categories, suggesting stronger text understanding and reasoning abilities.
While SDXL performs better with NNP, VBP, and WRB, which are often related to human-centric concepts.
This aligns with the known weakness of SD3 in body generation.

For numerical and chronological challenges, the CD (cardinal digit) category consistently ranks high in failure rates across all four models.
Most failures involve pure numbers or decade terms, indicating significant weaknesses in counting and chronological knowledge.

\textbf{Finding 3. The failure rate of PRP\$ (possessive pronoun) ranks high in all three models.}
PRP\$ show high failure rates across models. Words like ``its'', ``his'', and ``their'' rely heavily on context and lack specific referents, challenging the models' textual understanding capabilities. Similar issues arise with possessive endings (POS) and certain noun phrases, such as ``dog's owner'' and ``A strong bond of security''.
The reflections of DyEval also mention: ``The model seems to struggle when there is insufficient context provided in the text prompt.''

\input{Figures/sentence_length}
\input{Table/reflection_cases_abbreviated}
\textbf{Different types of failures.}
Based on our analysis and observation of the model failure data and inspired by multimodal hallucination~\cite{liu2023mitigating}, we categorize generation failures into three types: object error, relation error, and attribute error.
object error means the generated images contain objects that are poorly generated, missing, or additionally generated objects that are not mentioned in the input text and do not conform to human common sense.
Object error is that objects in generated images show poor visual quality; miss; or are not mentioned in the input text, and are harmonious with the overall image content.
Relation and attribute errors follow similar definitions.

We conducted hierarchical sampling of 10\% from failed text-image pairs, based on the topic and tested model.
We then manually annotated specific error types, calculated their frequency and grouped them by model and topic.
Two people annotate the same text-image pair and if the results are inconsistent, a third person is introduced.
As illustrated in Fig.~\ref{fig:error_analysy}, the advancement in model capabilities has led to a shift in bottleneck issues from objects to attributes to relations.
SD1-5 and SD2-1 predominantly fail in generating objects across most topics, while SDXL and SD3 show stronger object depiction capabilities, shifting difficulties to attribute generation.
SD3 exhibits fewer attribute errors than SDXL.

Interestingly, many test inputs in relation topics show higher rates for object error within images than for the relations themselves, also surpassing the rate of object error in the material object topic.
For example in Fig.~\ref{fig:error_analysis_case}, a test input for an implicit relation, ``A dog and its owner taking a leisurely walk in the park on a bright morning'', may show the man holding the dog's leash (indicating their relationship), yet the dog and man themselves are imperfectly generated.
This suggests the potential for two-stage generation approaches. Flawed objects could be redrawn to enhance overall generation quality.

\textbf{Length of test inputs.}
Similar to PoS analysis, we analyzed the lengths of all test inputs collected under the nine initial topics for the four models (excluding the contextual reflection module) and presented them in violin plots by depth, as shown in Fig.~\ref{fig:length}.
We can find test input length increases with the depth of DyEval, and failed test inputs are slightly longer than passed test inputs, suggesting that DyEval does not simply find model bugs by generating longer sentences.
Test inputs for SDXL and SD3 are slightly longer than those for SD1-5 and SD2-1, with the latter two models showing similar sentence length distributions.

\subsection{Intermediate process cases of DyEval}
\label{sec:ex_analysis}
In this section, we show that the contextual reflection module includes an interpretable dynamic failure location and bug analysis capability, which can further enhance the model in the future.

\textbf{Cases of dynamic failure location.}
Here we use two cases to show the dynamic failure location can find interpretable failure triggers of tested models in specific test inputs.
A case of SD3 as shown in Fig.~\ref{fig:miniInputs_case1}, firstly ``A luxurious silk kimono with elegant floral embroidery in vibrant hues.'' fails, and we split it and enter the next iteration.
The algorithm is executed until it reaches the smallest cell ``A kimono.'' and fails.
In that way, we can find the failure trigger in this test input is ``kimono'', and this also corresponds to the error-prone generation of SD3 for human body correlation in the previous section.

Fig.~\ref{fig:miniInputs_case2} shows another case of SD3, firstly the model generates ``A tiny black crescent moon within a big white cloud.''  failed. Then we split it into two sentences, and find they all pass.
According to our hypothesis, the limits of the capabilities of the tested models must be in their intersection sets.
Thirdly, we split ``A big white cloud.'' to ``A big cloud.'' and ``A white cloud.'', and merge them with ``A tiny black crescent moon.'' respectively, and they are all failed.
Following the algorithm, we continue to split ``A big cloud.'' and ``A white cloud.'' to ``A cloud'' by deleting attributes and merge with ``A tiny black crescent moon.'', and it still fails.
``A cloud'' reaches the smallest cell and the process ends.
This is a combinatorial error, ``A tiny black crescent
moon'' and ``within a cloud'' can not generate successfully in one test input, which can also be found in dynamic failure locations.

We provide the full process with details of the scene graph spliting and merging in the supplementary material.

\textbf{Cases of self-reflections.}
As shown in Table~\ref{tab:case_ablation_miniInputs_small}, LLM can indeed do some initial reflective summaries for the user assessments, and with the addition of dynamic failure location LLM's reflections are richer, longer, and more accurate in identifying mistakes.
The full version is in the supplementary material due to space constraints here.

We further conducted a survey among 10 domain experts specializing in text-to-image generation systems to evaluate the effectiveness of reflection contents in DyEval. Each expert was presented with one randomly selected reflection content from DyEval test nodes (an example can be found in the Reflections with dynamic failure location in Table~\ref{tab:case_ablation_miniInputs_small}). All experts (100\%) unanimously confirm that the reflections provide valuable and insightful content for model improvement and accurately correspond to the test records. Although some experts note the presence of redundant information, the majority (80\%) emphasize that the structured analysis offers actionable feedback for enhancing model performance.

%% file: Figures/Experiment1.tex
\begin{figure}[t]
  \centering
  \includegraphics[width=\linewidth]{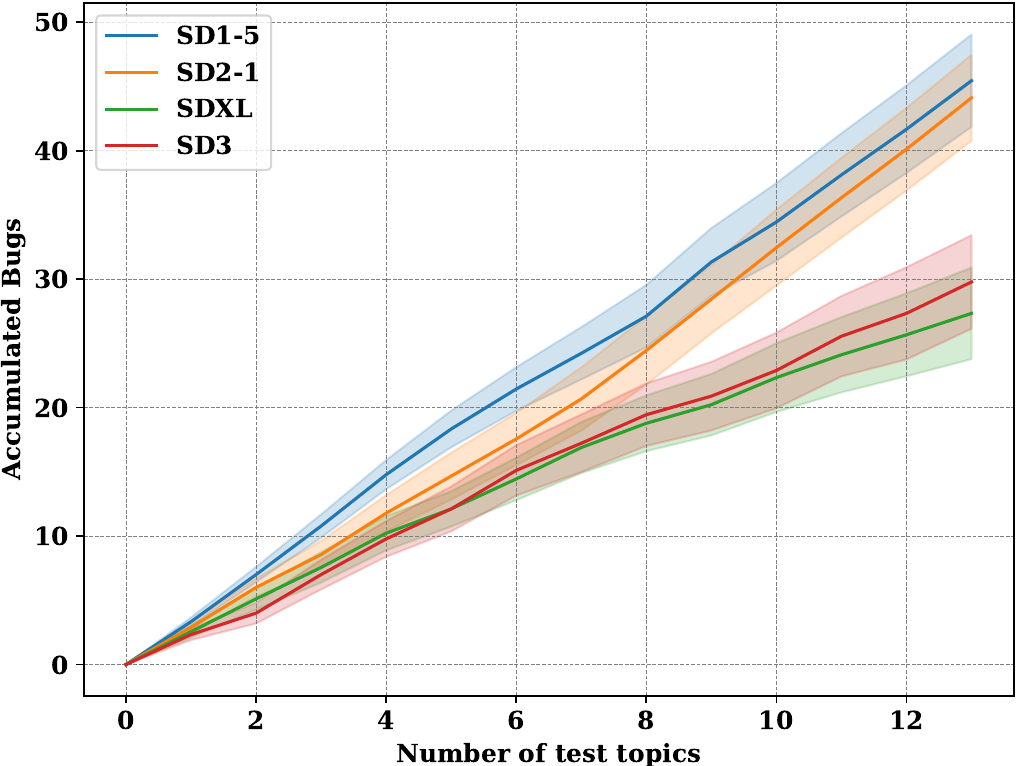} 
   \caption{Average number of bugs accumulated within an initial topic throughout thirteen test nodes (1 + 1$\times$3 + 3$\times$3) during the testing process of DyEval of SD1-5, SD2-1, SDXL, SD3.
    The standard error is over nine initial topics.
    The index of the test node of the test tree is obtained according to the breadth-first search.
    DyEval can constantly find bugs in the model under test, and weaker models are likelier to find bugs (SD1-5, SD2-1 consistently higher than SDXL, SD3). The shading represents the variance.} 
    \label{fig:exp2}
\end{figure}


%% file: Figures/Ablation1.tex



\begin{figure}
    \includegraphics[width=1\linewidth]{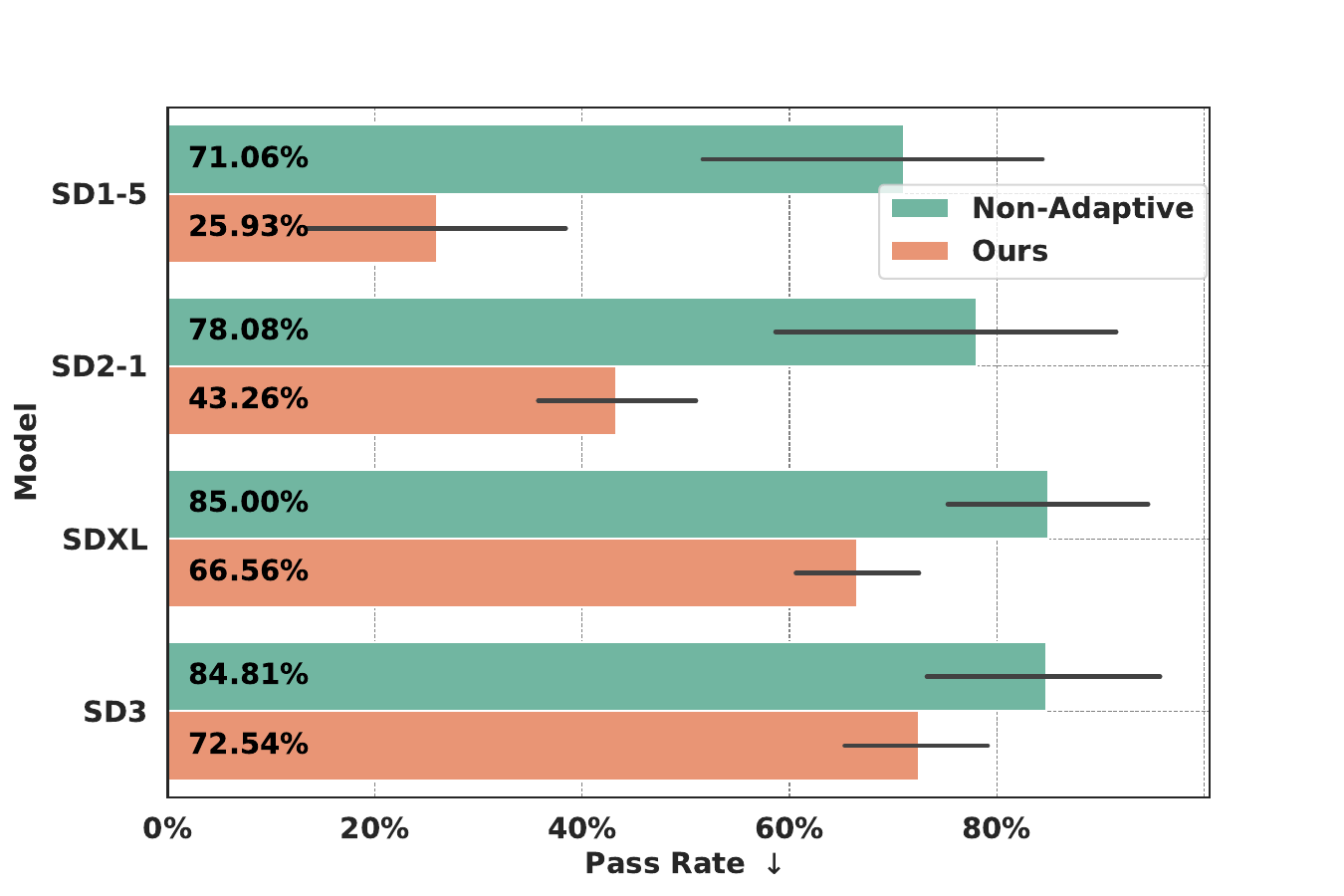}
    \caption{Comparison experiments between DyEval and non-adaptive testing (LLM directly generates the same amount of text input). The horizontal line represents the variance. Compared to the non-adaptive fashions, DyEval can find more generation failures in assessing the same number of text-image pairs, with greater distinguishability across models and better stability across test aspects.
    }
    \label{fig:ablation}

\end{figure}

%% file: Figures/Case.tex
\begin{figure}
    \centering
    \includegraphics[width=1\linewidth]{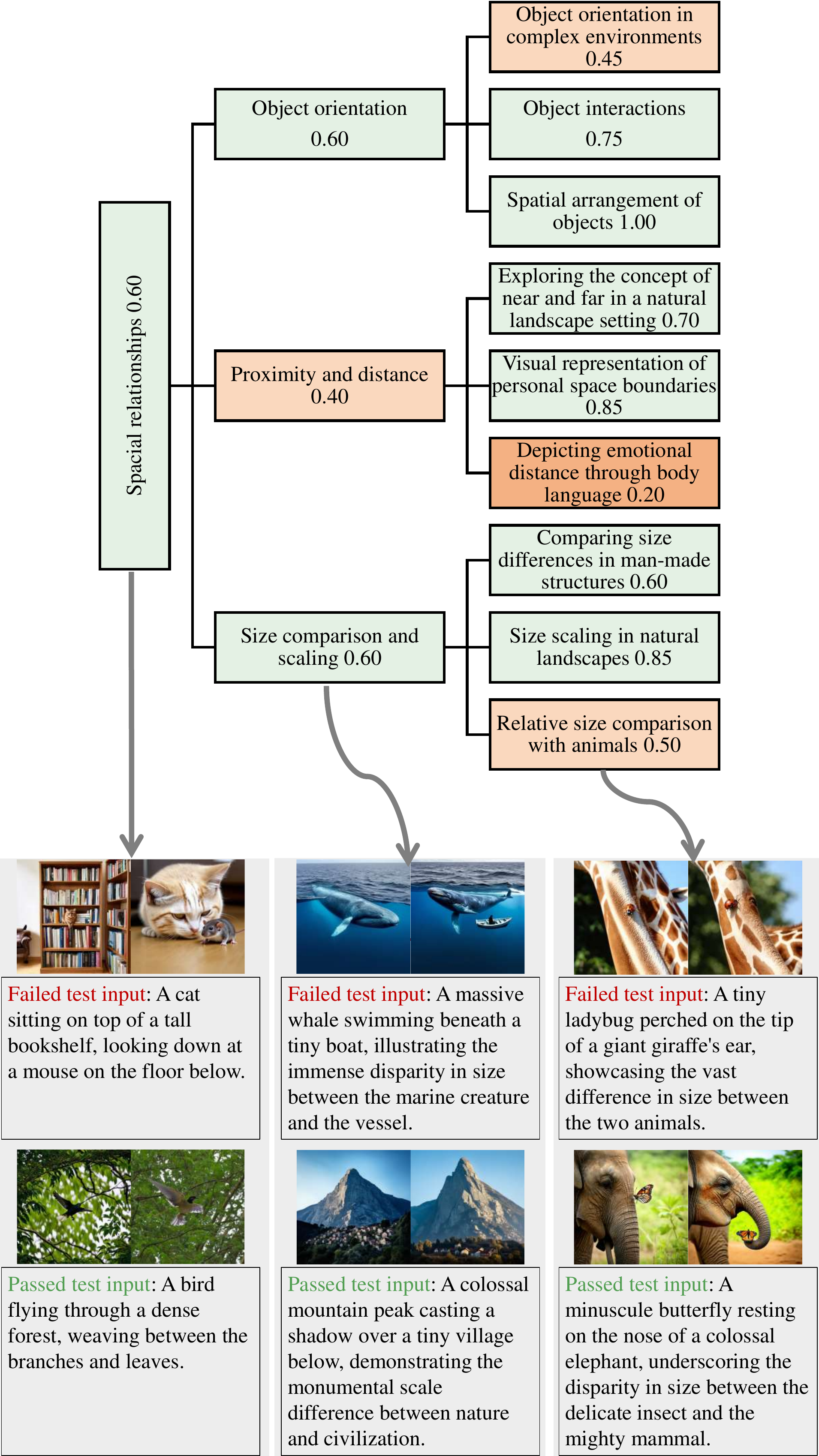}

    \caption{A test process case of DyEval. The initial test topic is ``Spacial relationships'' and the APR of this test node is 0.60. Based on the test records of the current topic, LLM continuously generates new test topics to analyze the model capability boundaries further.
    We also provide specific test inputs in the path of the test tree.
    Green represents nodes with test pass rate greater than or equal to 0.6, light orange represents pass rate (0.6,0.3], and less than 0.3 is dark orange.}
    \label{fig:case}
\end{figure}

%% file: Figures/ladar.tex
\begin{figure}[t]
   \includegraphics[width=\linewidth]{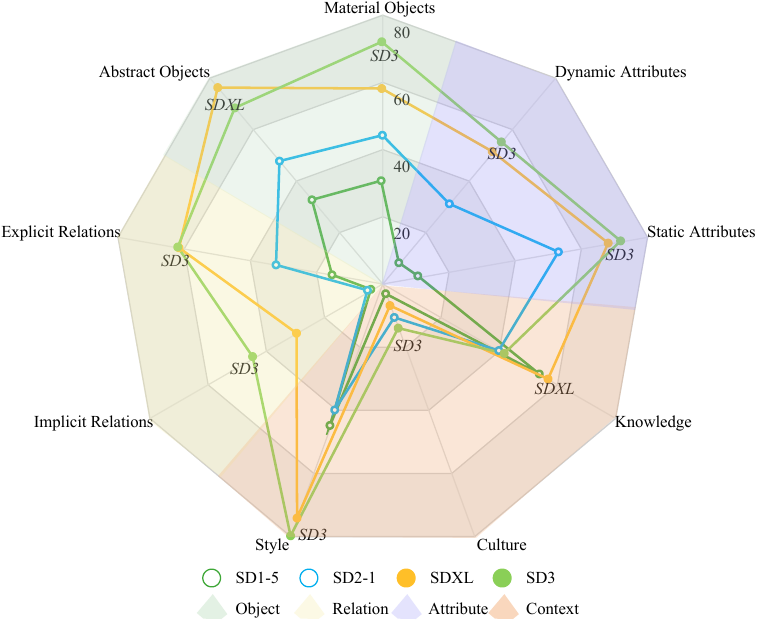} 
    \caption{Average test pass rate $APR$ of SD1-5, SD2-1, SDXL, and SD3 models in nine dimensions. Dimensions of the same base color indicate belonging to the same category. The overall trend of image generation capabilities is SD3>= SDXL> SD2-1> SD1-5; culture and implicit relations are the weakest part of all models.
    We label the best-performing models in different dimensions with the its name.}
    \label{fig:exp1}
\end{figure}

%% file: Figures/Experiment2.tex
\begin{figure*}[t]
    \centering
    \includegraphics[width=\linewidth]{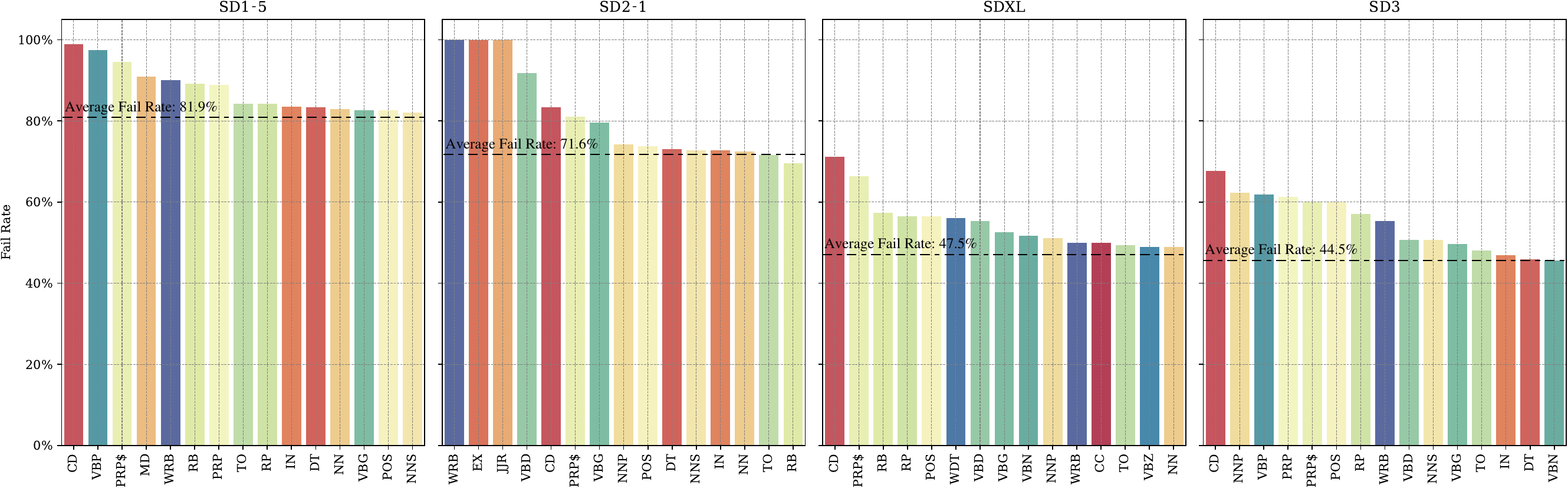}
    
    \caption{
    Top 15 error-prone parts-of-speech from different models, ordered by fail rate. The red dashed line represents the average fail rate for each part of speech in the model under test.
    }
    \label{fig:pos_tagging_analysiy}
\end{figure*}

%% file: Figures/Error_analysis.tex
\begin{figure*}[!t]
\centering
\includegraphics[width=0.24\textwidth]{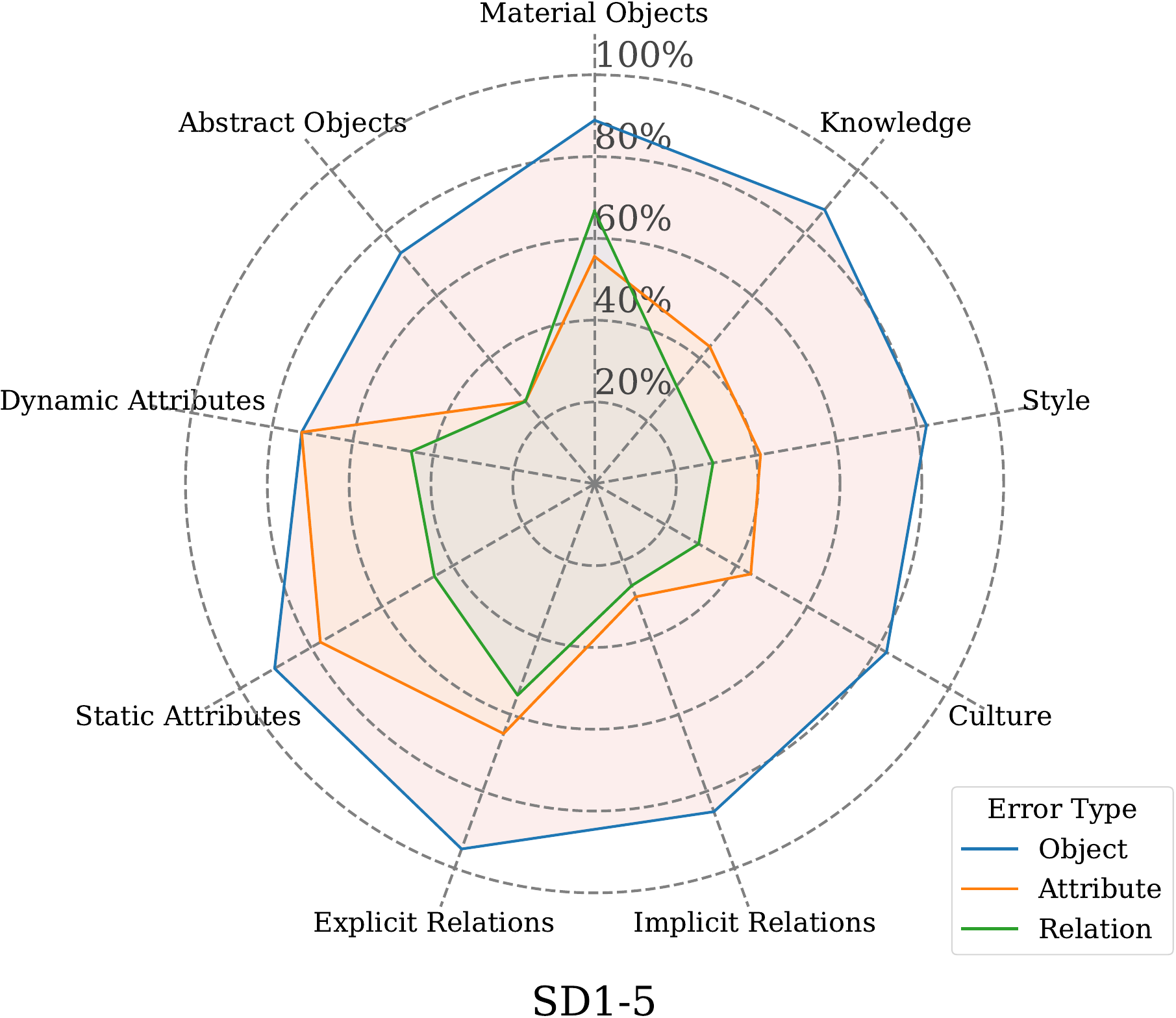}
    \label{fig:sd1_5}
\hfil
\includegraphics[width=0.24\textwidth]{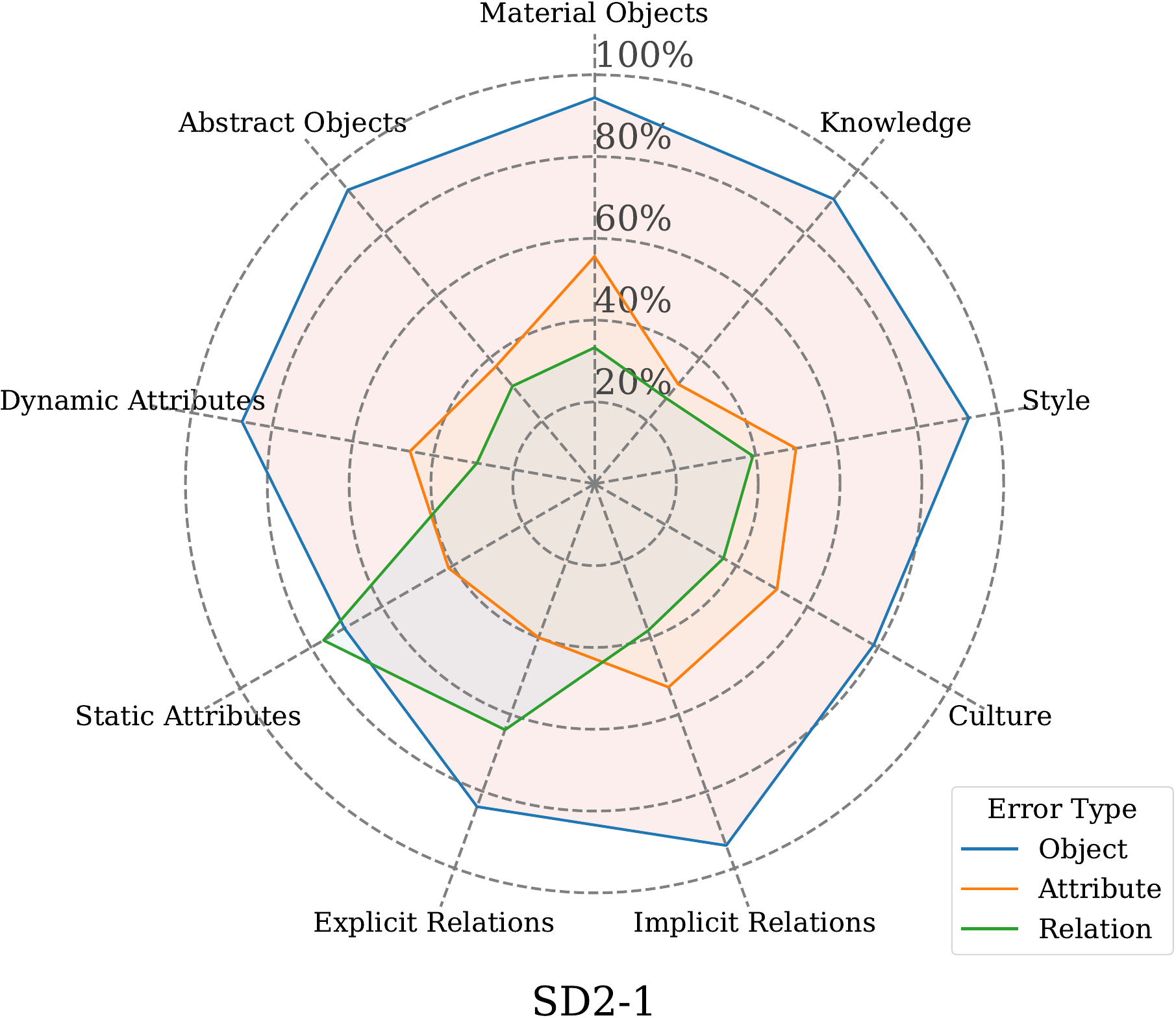}
    \label{fig:sd2_1}

\hfil

\includegraphics[width=0.24\textwidth]{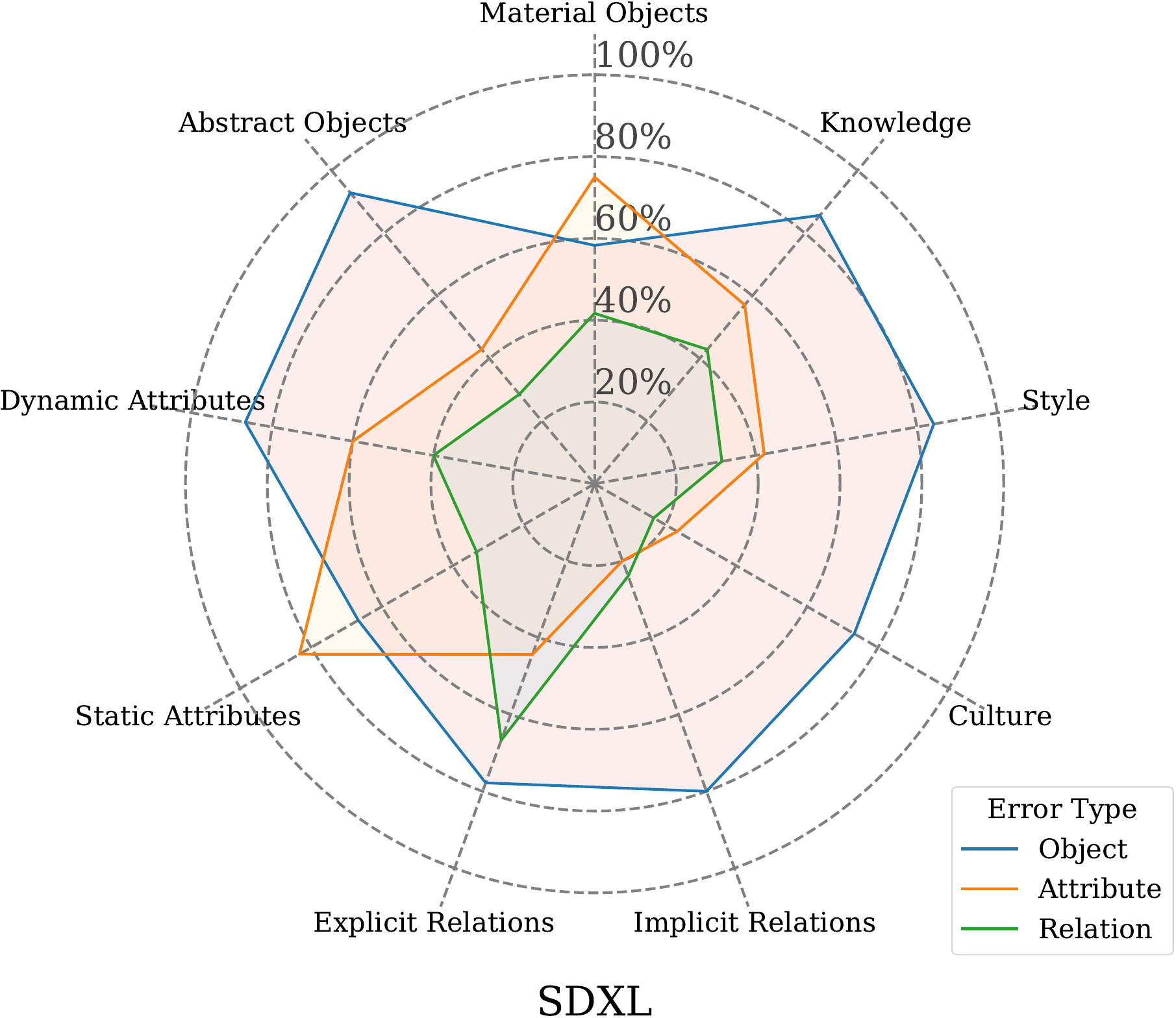}
    \label{fig:sdxl}
\hfil
\includegraphics[width=0.24\textwidth]{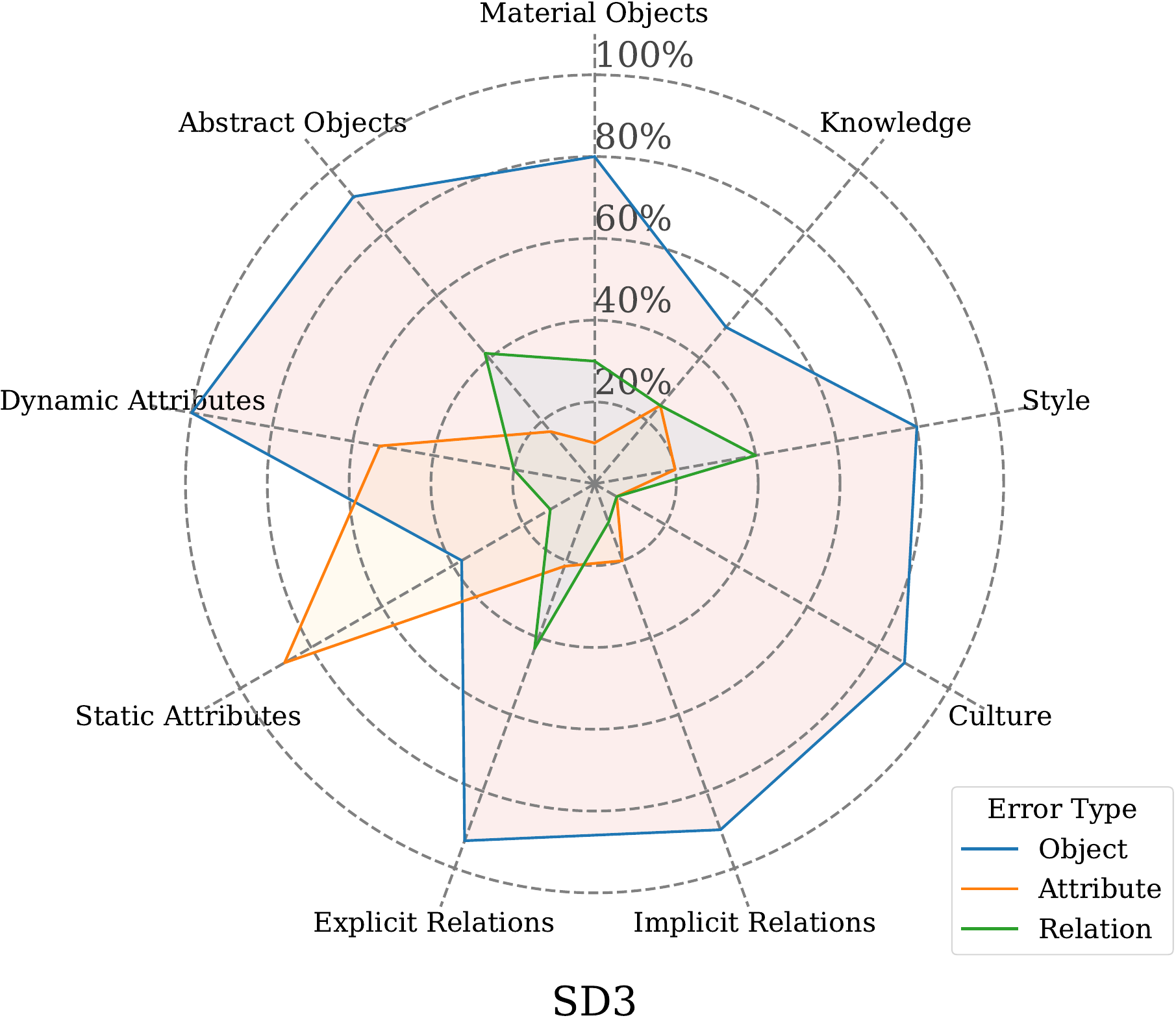}
    \label{fig:sd3}
\caption{Frequency of object, attribute, and relation errors in different models.}
  \label{fig:error_analysy}
\end{figure*}

%% file: Figures/error_analysis_Case.tex
\begin{figure}
    \centering

    \includegraphics[width=0.8\linewidth]{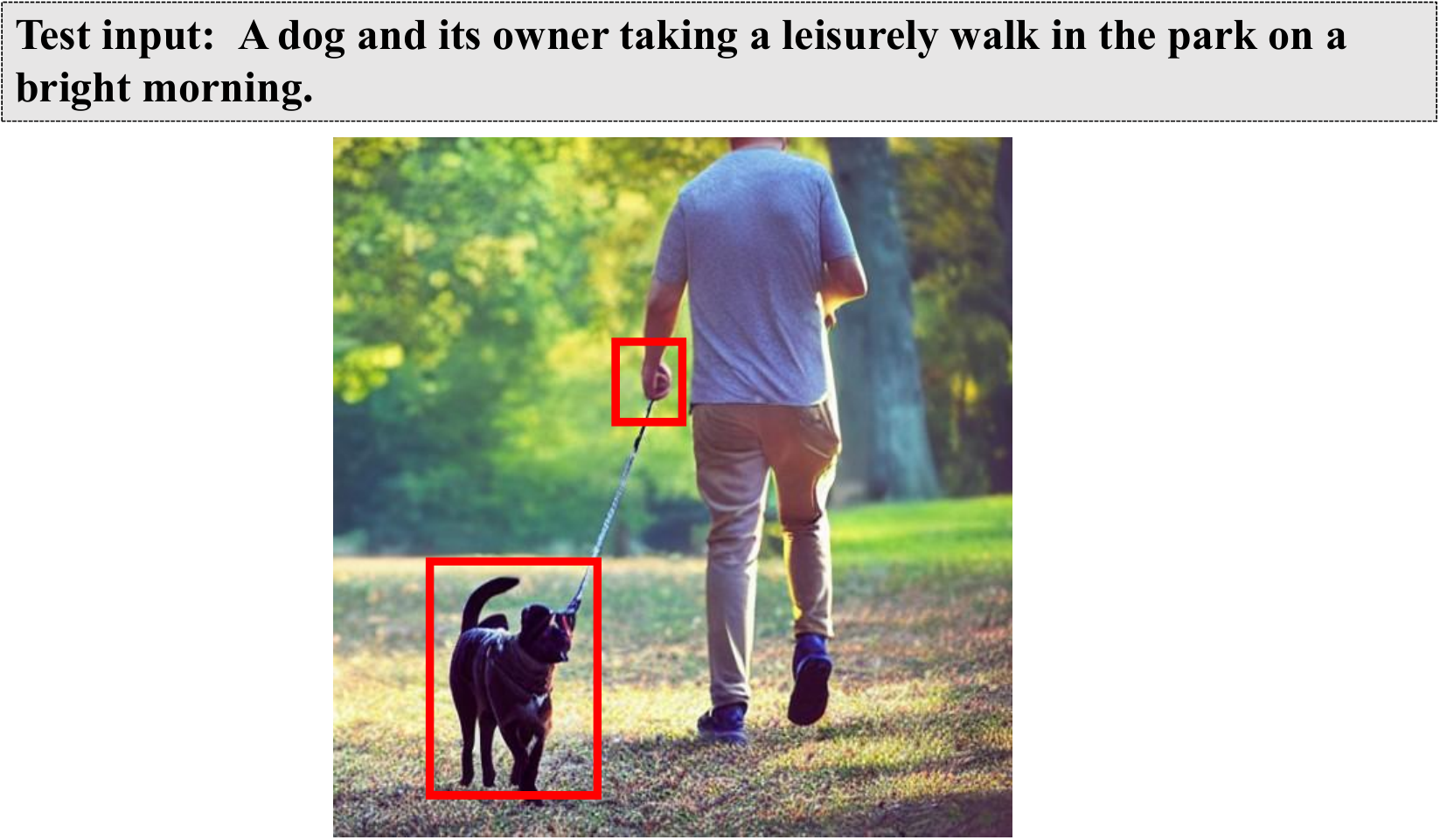}

    \caption{An error analysis example mentioned in the paper. Given the test input ``A dog and its owner taking a leisurely walk in the park on a bright morning.'' to test relation generation for SD1-5. While the model correctly generates a leash connecting two subjects, it fails to properly generate the dog and man themselves.
    } 
    \label{fig:error_analysis_case}
\end{figure}

%% file: Figures/sentence_length.tex
\begin{figure}[htbp]
  \centering
  \begin{subfigure}[b]{0.5\textwidth}
    \includegraphics[width=\textwidth]{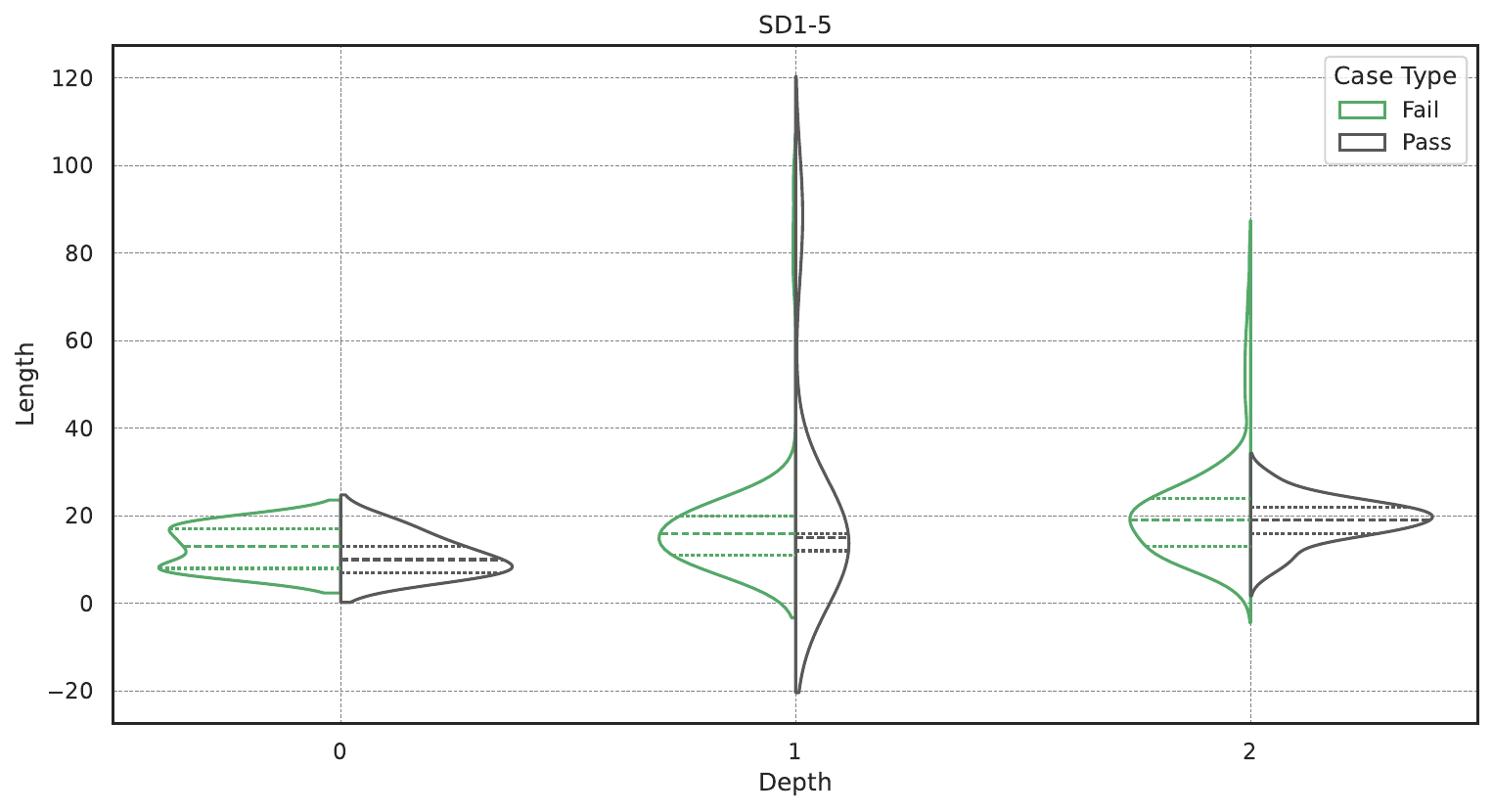}
    \caption{}
    \label{fig:sd1_5}
  \end{subfigure}
  \hfill 
  \begin{subfigure}[b]{0.5\textwidth}
    \includegraphics[width=\textwidth]{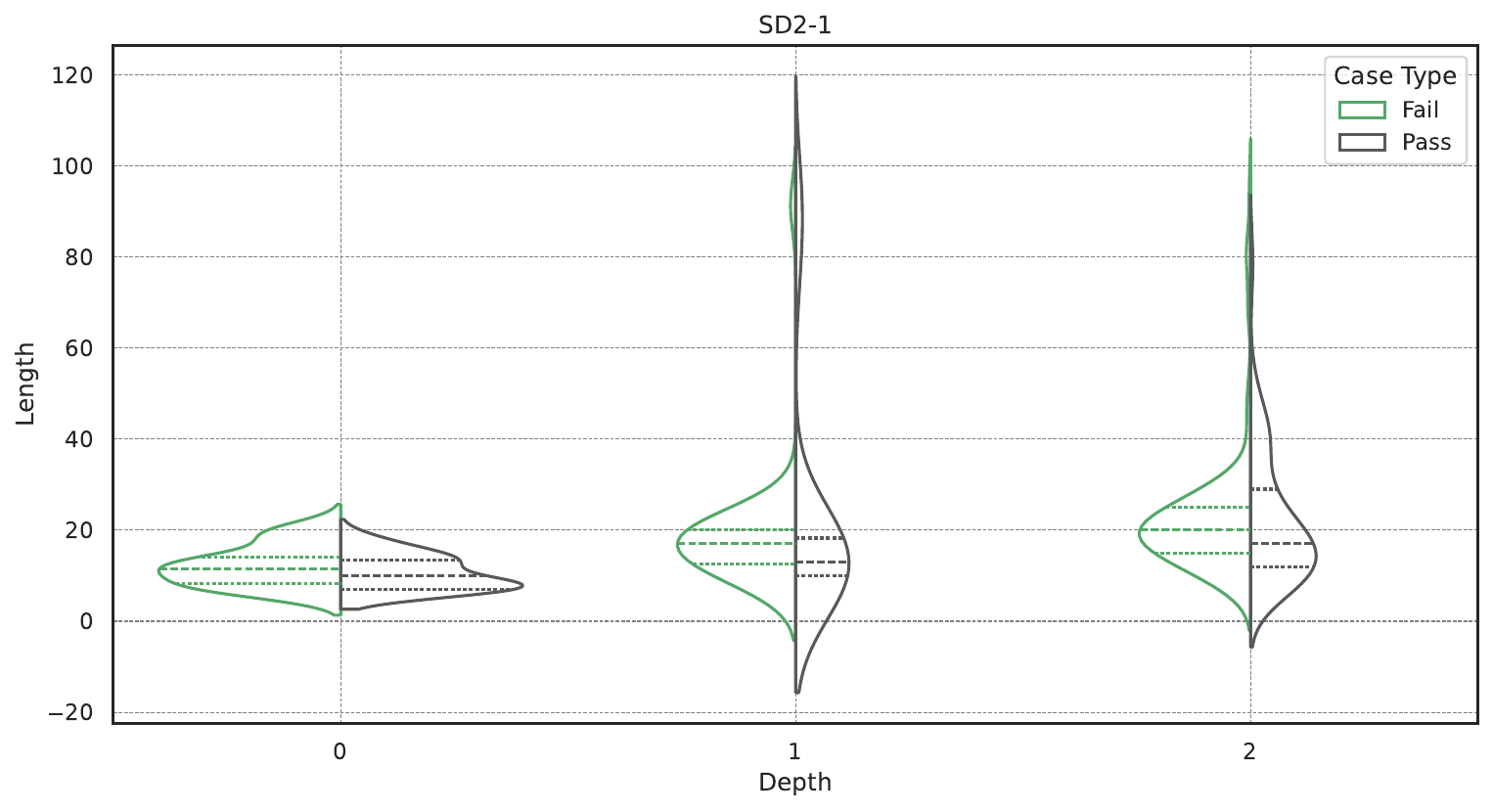}
    \caption{}
    \label{fig:sd2_1}
  \end{subfigure}
  \hfill 
  \begin{subfigure}[b]{0.5\textwidth}
    \includegraphics[width=\textwidth]{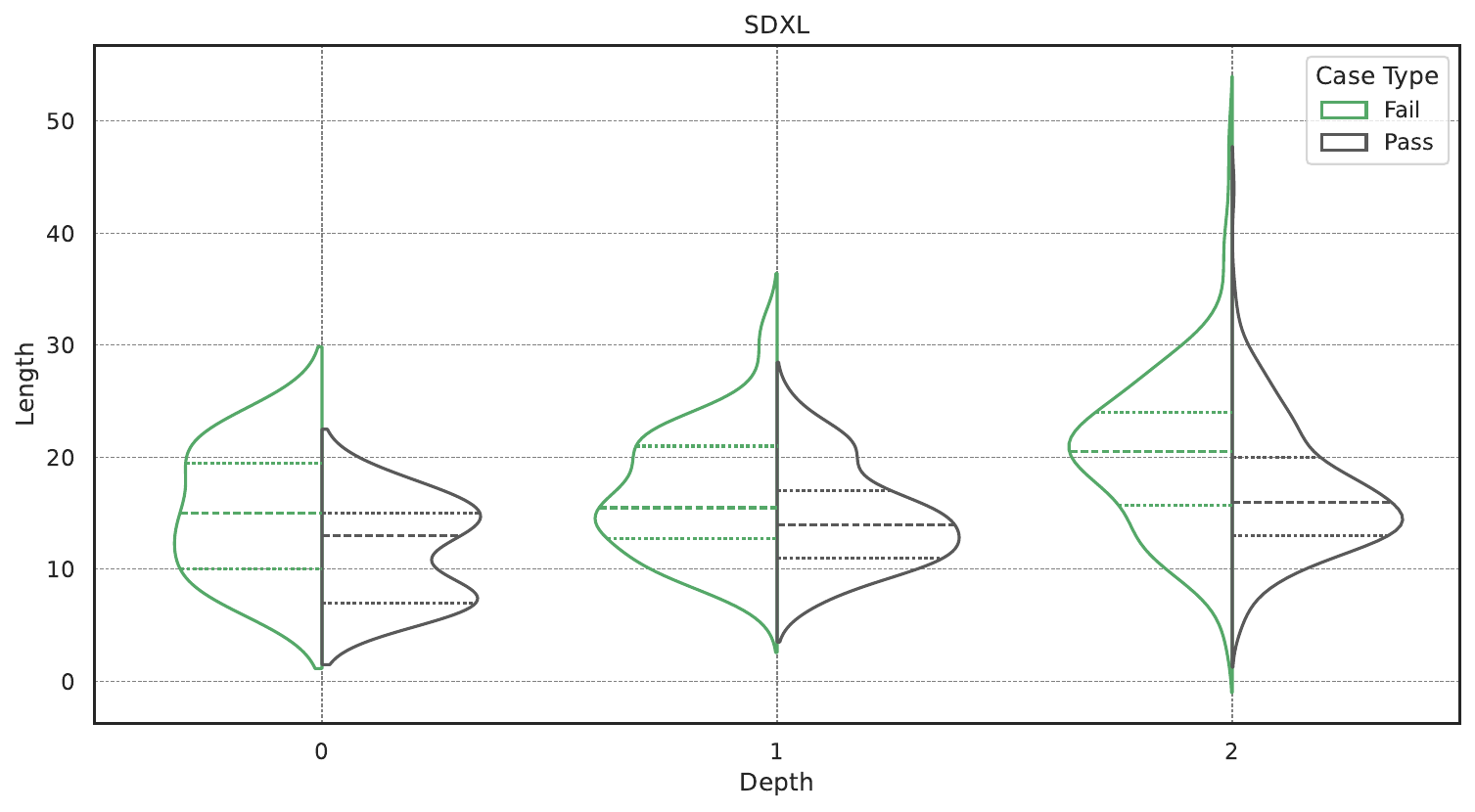}
    \caption{}
    \label{fig:sdxl}
  \end{subfigure}
  \begin{subfigure}[b]{0.5\textwidth}
    \includegraphics[width=\textwidth]{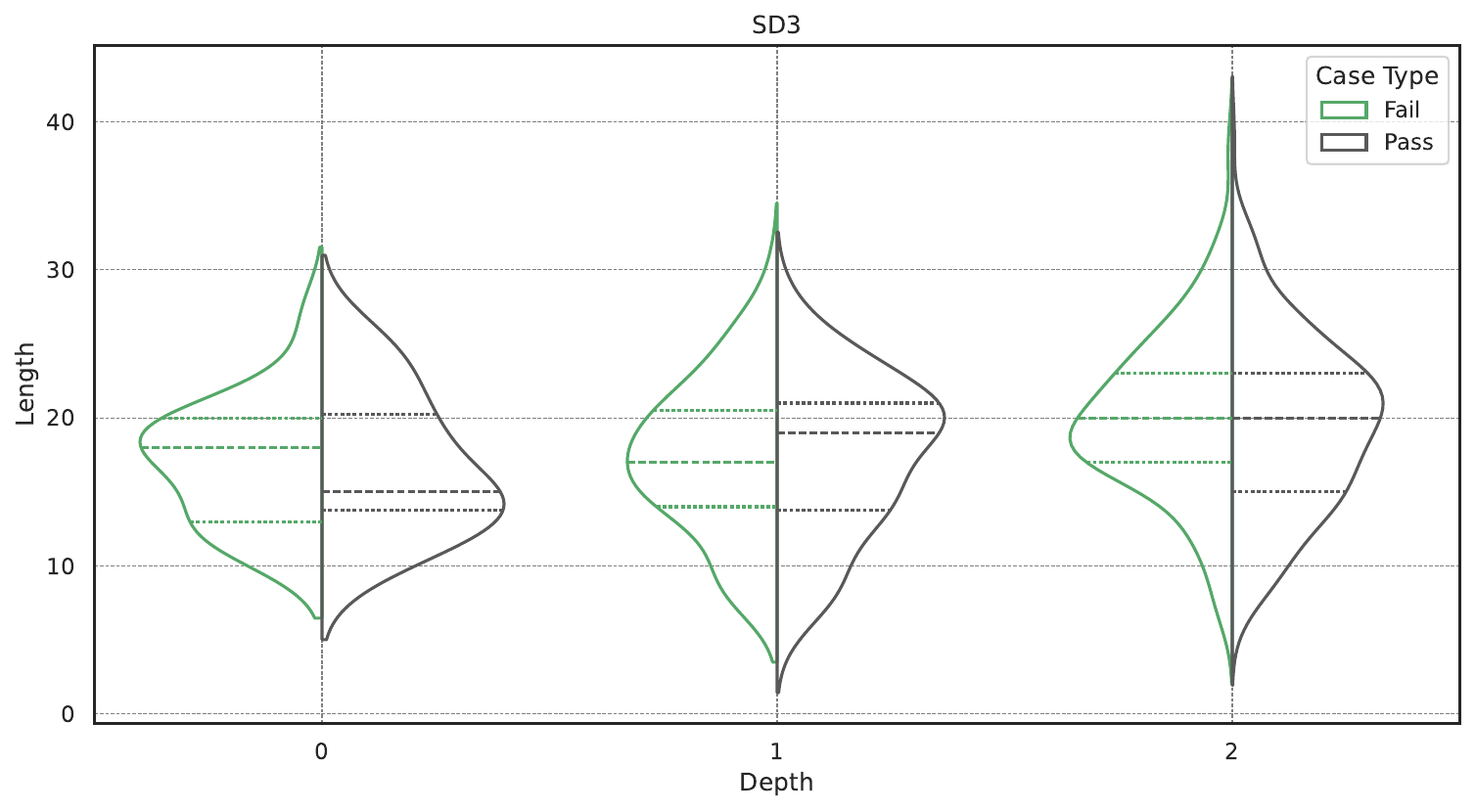}
    \caption{}
    \label{fig:sd3}
  \end{subfigure}
  \vspace{-1mm}
  \caption{Violin plot on the length of fail test inputs and pass test inputs in different depths across nine test records for different models. The three dashed lines in each subplot represent, from top to bottom, the upper quartile, median, and lower quartile, respectively.}
  \label{fig:length}
  \vspace{-7mm}
\end{figure}

%% file: Table/reflection_cases_abbreviated.tex
\begin{table*}[hbt!]
\caption{Case Study of reflections in DyEval, showing the influence of dynamic failure location and usefulness of contextual reflection module. Here 0 means fail and 1 means pass in ``Test inputs''. The textual expression is abbreviated here, the full version is in the supplementary material}
\label{tab:case_ablation_miniInputs_small} 
\centering
\begin{tabular}{p{0.15\linewidth}|p{0.8\linewidth}}
\toprule
\textbf{Type} & \textbf{Context}\\
\midrule
Test topic & Dog-human relationships \\
\midrule
Test inputs & 
\begin{itemize}
    \item A child hugging a dog tightly with a big smile on their face. 1
    \item A dog lying next to its owner on the couch, both fast asleep. 0
    \item A dog licking its owner's face to show affection and love. 0
    \item A dog cuddling up to its owner on a cold winter night, keeping them warm. 0
     \item A child hugging a dog tightly with a big smile on their face. 1
 \item A dog sitting patiently next to its owner as they read a book. 1
\item A dog jumping up to greet its owner with excitement as they come home. 1
\item A dog lying next to its owner on the couch, both fast asleep. 0 
\item A dog bringing its owner a toy, eagerly waiting for them to play. 1 
\item A dog licking its owner's face to show affection and love. 0 
\item A dog following its owner around the house, never wanting to be apart. 1 
\item A dog barking loudly to protect its owner from a potential threat. 1 
\item A dog cuddling up to its owner on a cold winter night, keeping them warm. 0
\end{itemize} \\
\midrule
Reflections w/o dynamic failure location &
\begin{itemize}
    \item Lack of Context: Model may fail due to insufficient context.
    \item Misinterpretation of Actions: Difficulty in understanding specific actions.
    \item Lack of Emotional Understanding: Struggles with representing emotions.
    \item Limited Understanding of Relationships: Challenges in capturing nuances of dog-human interactions.
\end{itemize} \\
\midrule
Reflections w/ dynamic failure location &
\begin{itemize}
    \item \textbf{Specific Actions}: Model struggles with complex interactions (e.g., "dog licking owner's face").
    \item \textcolor{red}{\textbf{Object Focus}}: Better performance with object-oriented inputs (e.g., "A dog's tail").
    \item \textbf{Contextual Relationships}: Difficulty in representing complex scenarios (e.g., "dog cuddling owner on cold night").
    \item \textbf{Incomplete Descriptions}: Surprisingly good performance with general concepts.
    \item \textcolor{red}{\textbf{Owner Focus}}: Consistent struggles when input focuses solely on the dog's owner.
\end{itemize} \\
\bottomrule
\end{tabular}
\end{table*}

%% file: Section/APPENDIX.tex
\newpage
\newpage
\appendix
\section{Experimental Details}
In our experiments, the LLM in our testing process sometimes generates more than the required number of test inputs or topics, at which time the participants will be selected, but sometimes there will be less than the required number so that some participants will do some more topics or inputs.
For data cleaning, we will clean some text-image pairs of the same content but opposite label results by the time spent on labeling filtering, generally short time is the user's misuse; and, we will filter out some of the test topics/inputs that do not be labeled.
Finally, we collect a total of 24,054 text-image pairs for the three models, of which there are 5,952 for SD1-5, 6,558 for SD2-1, 5,420 for SDXL, and 6,124 for SD3.

\section{Detailed Process of Dynamic Failure Location}
\subsection{Split function.}
We convert the text to a scene graph $c$ and then split it. The transformation between text and scene graph is done using LLM, see Table~\ref{tab:prompt_topic_input} for the specific task instruction prompts.

During the split and subsequent merge of the scene graph, the integrity of the dependencies must be maintained. This requires maintaining the association relations, attributes, and their corresponding objects. The segmentation strategy initially targets relation nodes; if a single relation exists or no relation exists, the focus shifts to the split of object and attribute nodes until atomic-level granularity is achieved ( only one node).

When merging different scene graphs, first put all the nodes on one graph, then determine if there are duplicate object and relationship nodes, and unify them.

As shown in Fig.~\ref{fig:scene_grapg_cases}, (a) shows a scene graph and (b) (c) are its sub-graph after splitting; while (b) (c) can merge, and the result is (a).
\input{Figures/scene_graph_case}

\section{Demo.}
Please see Fig.~\ref{fig:demo_page}.
\input{Figures/demos}

\section{Case Study}

\input{Figures/Case_dalle}

\paragraph{Cases of DALL$\cdot$E 3.}
As shown in Fig.~\ref{fig:case_dalle1} and Fig.~\ref{fig:case_dalle2}, we show four difficult and interesting test inputs found by Inf-Eval, and the corresponding results of DALL$\cdot$E 3, one of the SOTA text-to-image models that can now access, with the first three failing and the last succeeding, which echoes previous analyses that testing small models can help to find vulnerabilities in larger ones.
In Fig.~\ref{fig:case_dalle2}, we show specific test inputs and their topics generated by DyEval and tested on three models. dynamic color changes (case 1) and a complex blend of concrete and abstract objects (case 4).
This demonstrates that DyEval can be both diffuse and focused, proving its effectiveness as a test system. 

\subsection{Cases of reflections.}
As shown in Table~\ref{tab:case_ablation_miniInputs}, LLM can indeed do some initial reflective summaries for the user's reference, and with the addition of dynamic failure location LLM's reflections are richer, longer, and more accurate in identifying mistakes.
\input{Table/reflection_ablation_cases}

\section{More Details on POS Tagging Analysis Experiment}

\input{Table/pos_tagging_labels}
\input{Table/pos_tagging_words}
\input{Figures/pos_tagging_numbers}
We provide words with high failure rates in Table~\ref{tab:model_pos_tags_1_5}, Table~\ref{tab:model_pos_tags_2_1}, and Table~\ref{tab:model_pos_tags_xl}.
Table~\ref{tab:pos-tags} shows part-of-speech tags and their meanings.
We also show the number distributions of different parts of speech in different models in Fig.~\ref{fig:pos_tagging_number}, and the three models are basically consistent.

\input{Table/prompt}

\section{Task Prompts.}
Please see Table~\ref{tab:prompt_topic_input} and Table~\ref{tab:prompt_error_reflection}.

%% file: Figures/scene_graph_case.tex
\begin{figure*}
    \centering
    \includegraphics[width=0.8\linewidth]{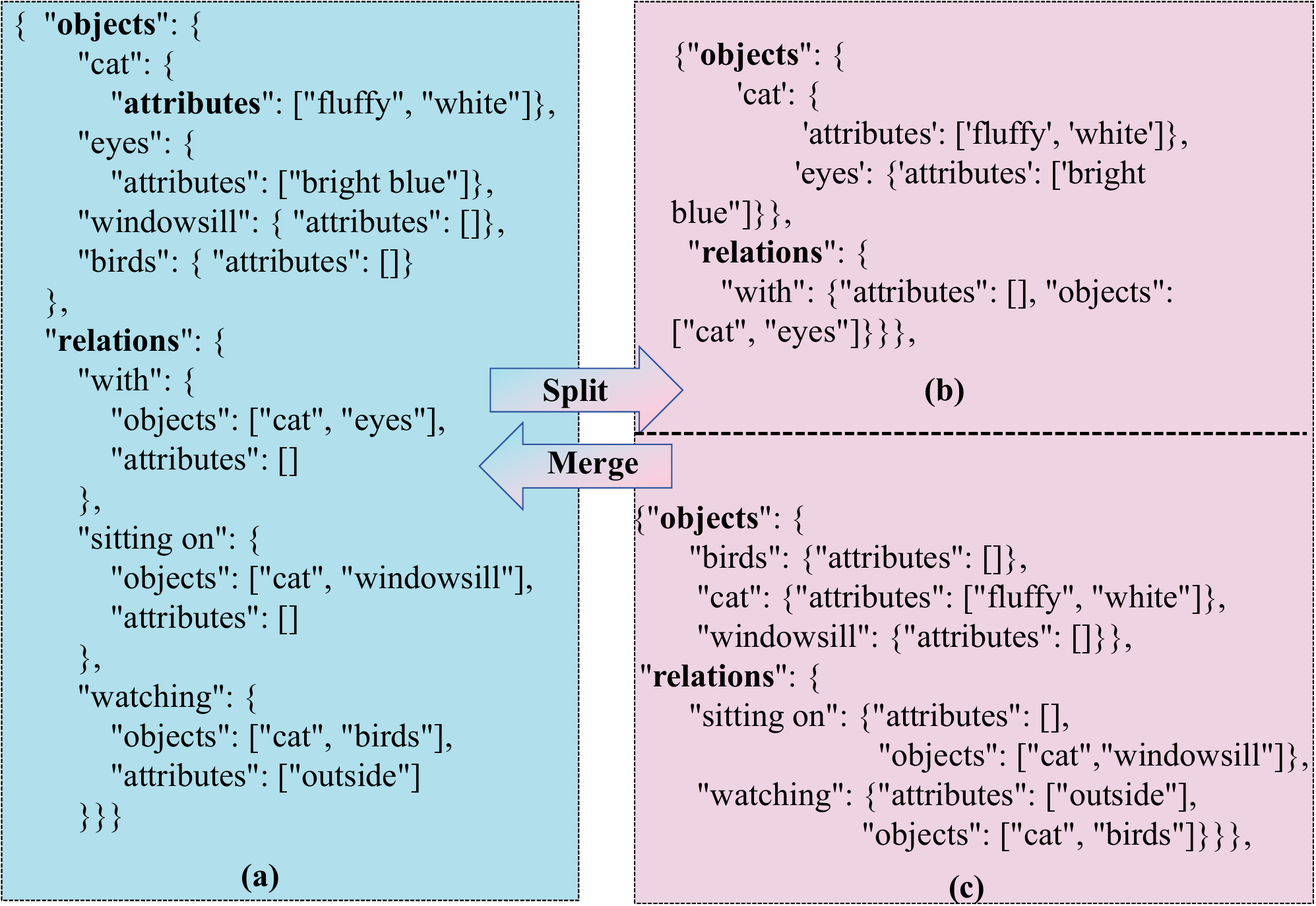}
    \caption{Example of split scene graph and merge scene graphs.
    }
    \label{fig:scene_grapg_cases}
\end{figure*}

%% file: Figures/demos.tex
\begin{figure*}[htbp]
  \centering

    \includegraphics[width=\textwidth]{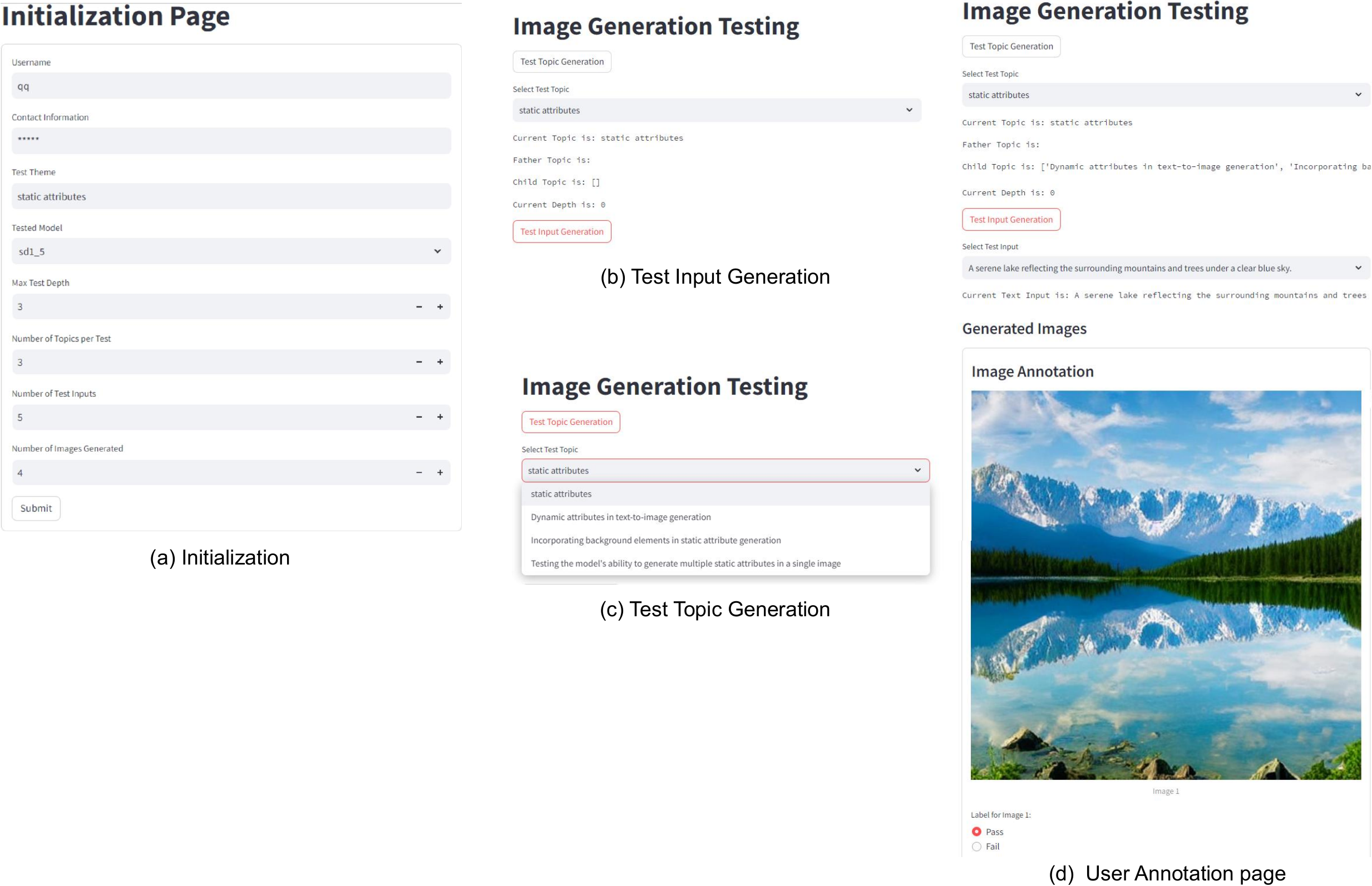}
    
  \caption{Demo system pages.}
  \label{fig:demo_page}
\end{figure*}

%% file: Figures/Case_dalle.tex
\begin{figure*}
    \centering
    \includegraphics[width=0.8\linewidth]{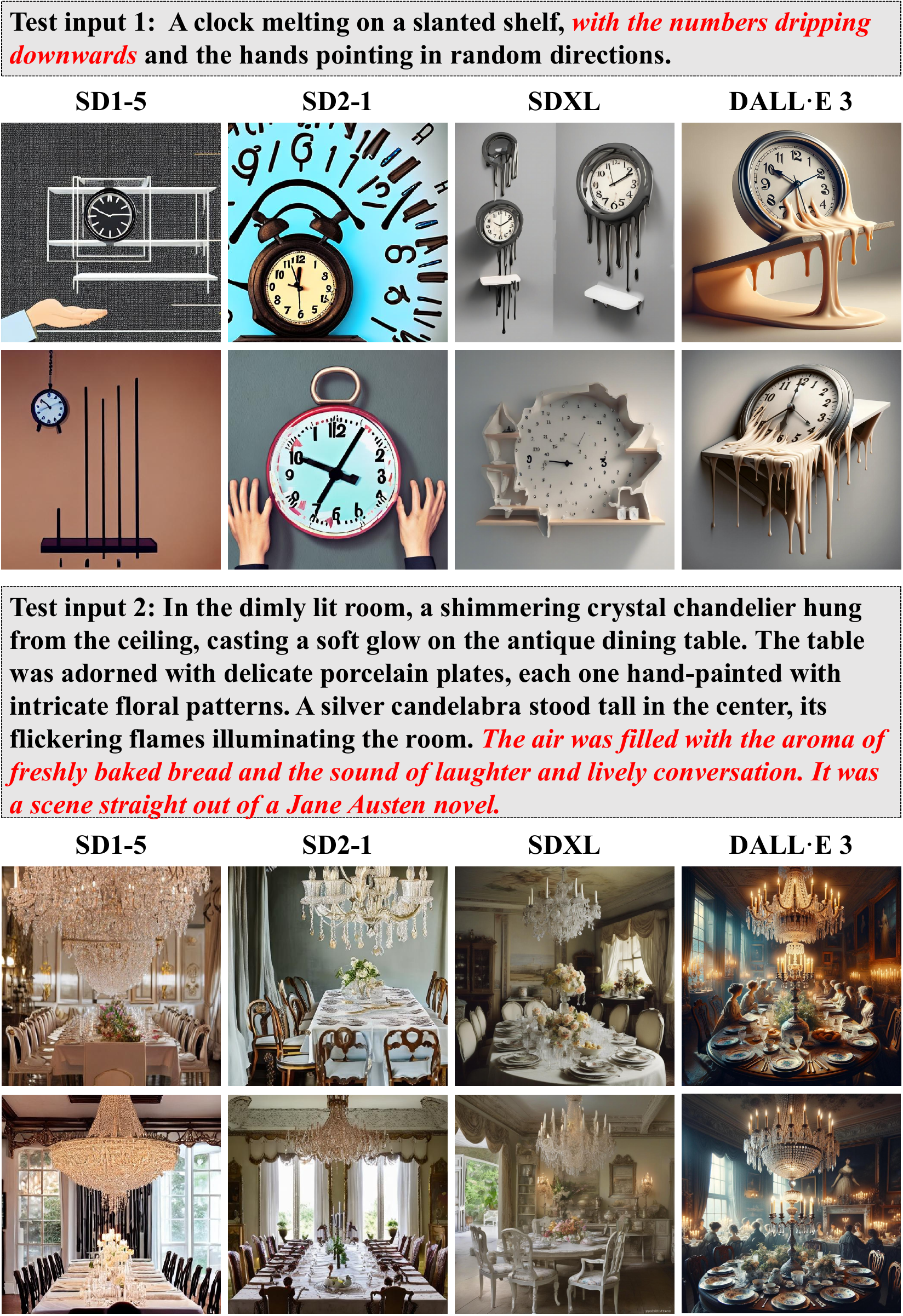}
    \caption{Interesting test inputs generated by DyEval. We find that all four models were poorly generated, with red indicating the missing parts of the four tested models.
    }
    \label{fig:case_dalle1}
\end{figure*}
\begin{figure*}
    \centering
    \includegraphics[width=0.8\linewidth]{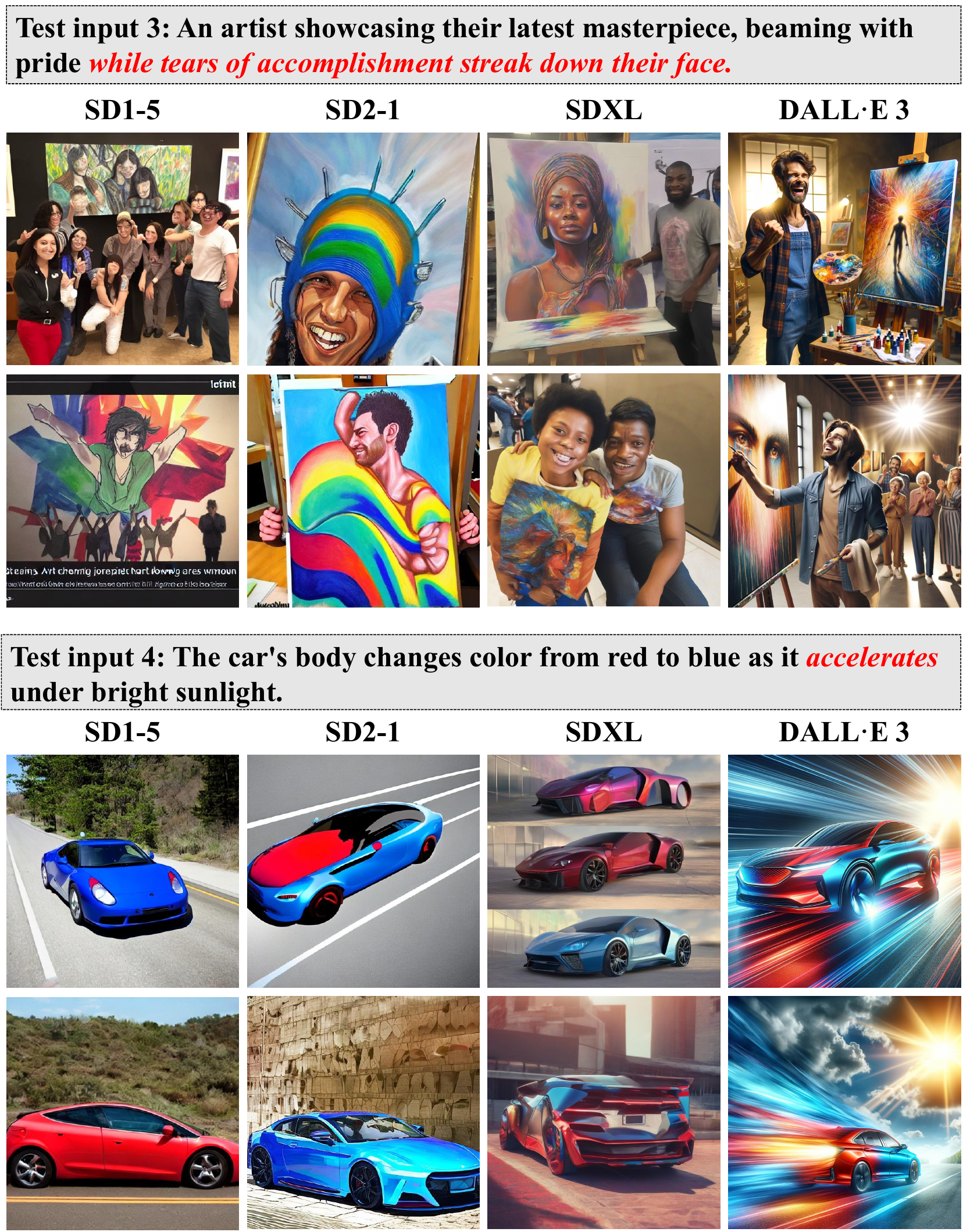}
    \caption{ Interesting test inputs generated by DyEval. The red color indicates the missing parts of the model under test, we find that in the first case, all four models are not generated well, the second case DALL·E 3 can generate satisfactory pictures.
    }
    \label{fig:case_dalle2}
\end{figure*}

%% file: Table/reflection_ablation_cases.tex
\newpage

\begin{table*}[hbt!]
\caption{Case Study of reflections in Inf-Eval. Here 0 means fail and 1 means pass in Test inputs. We show the influence of dynamic failure location in reflections.}
\label{tab:case_ablation_miniInputs} 
\centering
\begin{tabular}{p{0.105\linewidth}|p{0.845\linewidth}}

\toprule
\textbf{Type} & \textbf{Context}\\
\midrule

\multirow{1}{*}{Test topic} & Dog-human relationships \\
\midrule
\multirow{5}{*}{Test inputs} 
& A child hugging a dog tightly with a big smile on their face. 1 \\
& A dog sitting patiently next to its owner as they read a book. 1 \\
& A dog jumping up to greet its owner with excitement as they come home. 1 \\
& A dog lying next to its owner on the couch, both fast asleep. 0 \\
& A dog bringing its owner a toy, eagerly waiting for them to play. 1 \\
& A dog licking its owner's face to show affection and love. 0 \\
& A dog following its owner around the house, never wanting to be apart. 1 \\
& A dog barking loudly to protect its owner from a potential threat. 1 \\
& A dog cuddling up to its owner on a cold winter night, keeping them warm. 0 \\

\midrule
\multirow{5}{*}{\parbox{2.4cm}{Reflections w/o\\ dynamic failure\\ location}}  &
Failure Patterns:
1. Lack of Context: The model may fail to generate accurate images because it lacks the necessary context. In the given test records, the model may not understand the specific actions or emotions associated with the dog-human relationship, leading to incorrect or irrelevant image generation.\\

&2. Misinterpretation of Actions: The model may misinterpret the actions described in the text prompt, resulting in inaccurate image generation. For example, in the first test record, the model may not understand the concept of a dog wagging its tail while its owner scratches its belly, leading to a failure in generating a relevant image.\\

&3. Lack of Emotional Understanding: The model may struggle to understand and represent emotions accurately. In the third test record, where a dog is described as licking its owner's face to show affection and love, the model may not be able to capture the emotional aspect, resulting in a failure.\\

&4. Limited Understanding of Relationships: The model may have a limited understanding of the dynamics and nuances of dog-human relationships. In the fourth test record, where a dog is described as cuddling up to its owner on a cold winter night, the model may not be able to accurately represent the warmth and comfort associated with such a relationship, leading to a failure.\\

&Overall, the failure patterns suggest that the model may struggle with understanding context, actions, emotions, and the intricacies of dog-human relationships. These limitations can result in inaccurate or irrelevant image generation. \\ \midrule

\multirow{6}{*}{\parbox{2.8cm}{Reflections w/ \\ dynamic failure\\ location}}  &
Based on the provided test records, we can observe the following patterns: 
1. **Specific Actions**: The model seems to struggle when the test input includes specific actions or interactions between the dog and its owner. For example, in the test inputs "A dog wagging its tail while its owner scratches its belly" and "A dog licking its owner's face to show affection and love," the model fails to generate the corresponding images. It appears that the model may have difficulty understanding and representing complex actions or interactions. \\

&2. \textcolor{red}{**Object Focus**}: The model tends to perform better when the test input focuses on a specific object related to the dog, such as "A dog's tail" or "A dog lying on the couch." In these cases, the model successfully generates the corresponding images. This suggests that the model may be more proficient at generating images when the input is more object-oriented and less context-dependent. \\

&3. **Contextual Relationships**: The model struggles when the test input involves contextual relationships between the dog and its owner, such as "A dog lying next to its owner on the couch, both fast asleep" or "A dog cuddling up to its owner on a cold winter night, keeping them warm." These inputs result in failures, indicating that the model may have difficulty capturing and representing complex relationships or scenarios involving the dog and its owner. \\

&4. **Incomplete Descriptions**: In some cases, the test inputs provide incomplete descriptions, such as "A dog" or "A dog on a cold winter night." Surprisingly, the model successfully generates the corresponding images in these cases. It suggests that the model may rely more on the general concept of a dog or the given context (e.g., cold winter night) rather than specific details or actions. \\

&5. \textcolor{red}{**Owner Focus**}: The model consistently struggles when the test input focuses solely on the dog's owner, such as "A dog's owner" or "A dog's owner on the couch." These inputs consistently result in failures, indicating that the model may have difficulty generating images based solely on the owner's presence or actions.\\

&In summary, the failure patterns observed in the test records suggest that the text-to-image model may underperform in scenarios involving specific actions or interactions, contextual relationships, and descriptions solely focused on the dog's owner. The model seems to perform better when the input is more object-oriented, less context-dependent, and provides a complete description of the dog or the given context.\\ \bottomrule

\end{tabular}

\end{table*}

%% file: Table/pos_tagging_labels.tex
\begin{table*}[h!]
\centering

\begin{tabular}{ll}
\toprule
\textbf{Tag} & \textbf{Description} \\
\midrule

CC & Coordinating conjunction \\
CD & Cardinal number \\
DT & Determiner \\
EX & Existential there \\
FW & Foreign word \\
IN & Preposition or subordinating conjunction \\
JJ & Adjective \\
JJR & Adjective, comparative \\
JJS & Adjective, superlative \\
LS & List item marker \\
MD & Modal \\
NN & Noun, singular or mass \\
NNS & Noun, plural \\
NNP & Proper noun, singular \\
NNPS & Proper noun, plural \\
PDT & Predeterminer \\
POS & Possessive ending \\
PRP & Personal pronoun \\
PRP\$ & Possessive pronoun \\
RB & Adverb \\
RBR & Adverb, comparative \\
RBS & Adverb, superlative \\
RP & Particle \\
SYM & Symbol \\
TO & to \\
UH & Interjection \\
VB & Verb, base form \\
VBD & Verb, past tense \\
VBG & Verb, gerund or present participle \\
VBN & Verb, past participle \\
VBP & Verb, non-3rd person singular present \\
VBZ & Verb, 3rd person singular present \\
WDT & Wh-determiner \\
WP & Wh-pronoun \\
WP\$ & Possessive wh-pronoun \\
WRB & Wh-adverb \\

\bottomrule
\end{tabular}
\caption{Part-of-speech tags and their meanings.}
\label{tab:pos-tags}
\end{table*}

%% file: Table/pos_tagging_words.tex
\begin{table*}[htbp]
\centering
\begin{tabular}{l|l|p{8cm}}
\toprule
\textbf{Model} & \textbf{Part-Of-Speech tag}& \textbf{Error-prone words}\\
\midrule
\multirow{15}{*}{SD1-5} & CD & one, 2, 4, 30, 8, 10, two, 3, 6, 20, 1920s, 15, 7, 5, 1970s \\
& VBP & believe, color, face, are, use, guide, determine, artwork, argue, reflect, exist, see, worn, consist \\
& PRP\$ & their, her, its, his, our \\
& MD & will, may, can, should \\
& WRB & when, how \\
& RB & closely, then, partially, smoothly, also, however, together, around, lively, often, completely, not, gracefully \\
& PRP & them, it, us, she, they, we \\
& TO & to \\
& RP & off, over, up \\
& IN & in, with, that, on, around, from, under, as, into, of, for, while \\
& DT & any, the, some, each, a, all, these, no, this \\
& NN & concept, color, student, group, person, teacher, fashion, induction, orange, ball, dog, length \\
& VBG & making, transitioning, wearing, playing, holding, flying, consisting, overlapping, performing \\
& POS & 's \\
& NNS & gowns, questions, dresses, universals, trends, women, forms, dancers, centimeters, metaphysics \\
\bottomrule
\end{tabular}
\caption{Model SD1-5: The top fifteen high-fail lexical labels and error-prone words in each model, where error-prone words are those whose fail rate exceeds the average model fail rate.}
\label{tab:model_pos_tags_1_5}
\end{table*}

\begin{table*}[htbp]
\centering
\begin{tabular}{l|l|p{8cm}}
\toprule
\textbf{Model} & \textbf{Part-Of-Speech tag}& \textbf{Error-prone words}\\
\midrule
\multirow{15}{*}{SD2-1} & WRB & where, how \\
& EX & there \\
& JJR & older, better, flower, stronger, more, feather, soldier \\
& VBD & stood, covered, were, had, danced, was, adorned, rose, possessed, beaded, illuminated, dared, light, ripped \\
& CD & 1588, one, 1960s, two, 2000s, 1950s, 1815, 1773, 1980s, 1990s, 1970s, 1989, 1920s, 90 \\
& PRP\$ & their, her, his \\
& VBG & casting, creating, hanging, wearing, lying, playing, rolling, holding, dancing, featuring, overlapping, showcasing, performing \\
& NNP & deco, a, art, east, rome, colosseum, berlin, war, generate, charleston, bauhaus, wall, year, renaissance \\
& POS & 's \\
& DT & any, every, some, an, a, all, those, both \\
& NNS & elements, feathers, trees, streets, stars, decorations, dreams, beads, dancers, skyscrapers \\
& IN & with, over, towards, by, on, through, around, from, under, at, as, for, while \\
& NN & book, background, student, group, teacher, night, user, ball, city, image \\
& TO & to \\
& RB & sphere, gracefully, partially, smoothly, proudly, together, seamlessly, yet, lively, carefully, peacefully, attentively, parallel, tightly \\

\bottomrule
\end{tabular}
\caption{Model SD2-1: The top fifteen high-fail lexical labels and error-prone words in each model, where error-prone words are those whose fail rate exceeds the average model fail rate.}
\label{tab:model_pos_tags_2_1}
\end{table*}

\begin{table*}[htbp]
\centering
\begin{tabular}{l|l|p{8cm}}
\toprule
\textbf{Model} & \textbf{Part-Of-Speech tag}& \textbf{Error-prone words}\\
\midrule
\multirow{15}{*}{SDXL}& CD & 1960s, 2, 1990, two, 2000s, three, 1940s, 1980s, 1776, 5, 4, 8, 1920s, 7 \\
& PRP\$ & your, their, our, its, his \\
& RB & down, softly, endlessly, closely, beautifully, gracefully, slowly, together, yet, away, lively, infinitely, effortlessly, peacefully, tightly \\
& POS & ', 's \\
& RP & down, up, out \\
& WDT & that \\
& VBD & delivered, packed, recycled, draped, was, adorned, parked, stacked, curled, closed, filled, grilled, dressed, painted \\
& VBG & ensuring, depicting, wearing, symbolizing, flying, dancing, enjoying, representing, overlapping, showing, showcasing, capturing, performing \\
& VBN & represented, perched, gathered, based, woven, shared, paired, made, embellished, filled, surrounded, snuggled, dressed, worn \\
& NNP & diwali, festival, a, polaroid, india, york, venice, city, charleston, brazil, year, renaissance, italy, new \\
& WRB & when \\
& CC & or, but, and \\
& TO & to \\
& VBZ & dances, pens, fades, features, plays, accelerates, includes, slips, .., blends, emits, lingers \\
& NN & painting, photograph, background, dress, owner, group, illustration, fashion, beach, roof, ball, moment, family, safety \\

\bottomrule
\end{tabular}
\caption{Model SDXL: The top fifteen high-fail lexical labels and error-prone words in each model, where error-prone words are those whose fail rate exceeds the average model fail rate.}
\label{tab:model_pos_tags_xl}
\end{table*}

\begin{table*}[htbp]
\centering
\begin{tabular}{l|l|p{8cm}}
\toprule
\textbf{Model} & \textbf{Part-Of-Speech tag}& \textbf{Error-prone words}\\
\midrule
\multirow{15}{*}{SD3}& CD & four, 44, 1914, 1948, 1969, two, three, 1903, one, 1773, 1963, 1989, 1920s, 1865, 1776 \\
& NNP & york, rome, colosseum, great, war, roman, new, ai, revolutionary, mayan, battle \\
& VBP & dive, assist, chase, commute, acorns, rush, towards, explore, endure, move, are, amidst, sway, spike, fall \\
& PRP & him, he, them, they, i \\
& PRP\$ & his, her, their, our \\
& POS & 's, ' \\
& RP & out, up, off \\
& WRB & when, how, where \\
& NNS & crops, lights, people, trees, streets, musicians, dancers, warriors, animals, moments, hailstones, carvings, hands \\
& VBD & labeled, textured, sunlight, adorned, filled, light, shot, placed, made, dressed, fragmented, parked, faded, polished \\
& VBG & swirling, showing, sitting, performing, playing, representing, flying, being, carrying, symbolizing, featuring \\
& TO & to \\
& IN & from, under, in, during, around, at, as, into, by, for \\
& JJ & bustling, intricate, ancient, american, different, tiny, egyptian, traditional, japanese, modern, chinese \\
& VBN & surrounded, adorned, shaped, covered, left, nestled, shared, lost, inspired, depicted, perched, lined, integrated, focused \\
\bottomrule
\end{tabular}
\caption{Model SD3: The top fifteen high-fail lexical labels and error-prone words in each model, where error-prone words are those whose fail rate exceeds the average model fail rate.}
\label{tab:model_pos_tags_sd3}
\end{table*}

%% file: Figures/pos_tagging_numbers.tex
\begin{figure*}
    \centering
    \includegraphics[width=1\linewidth]{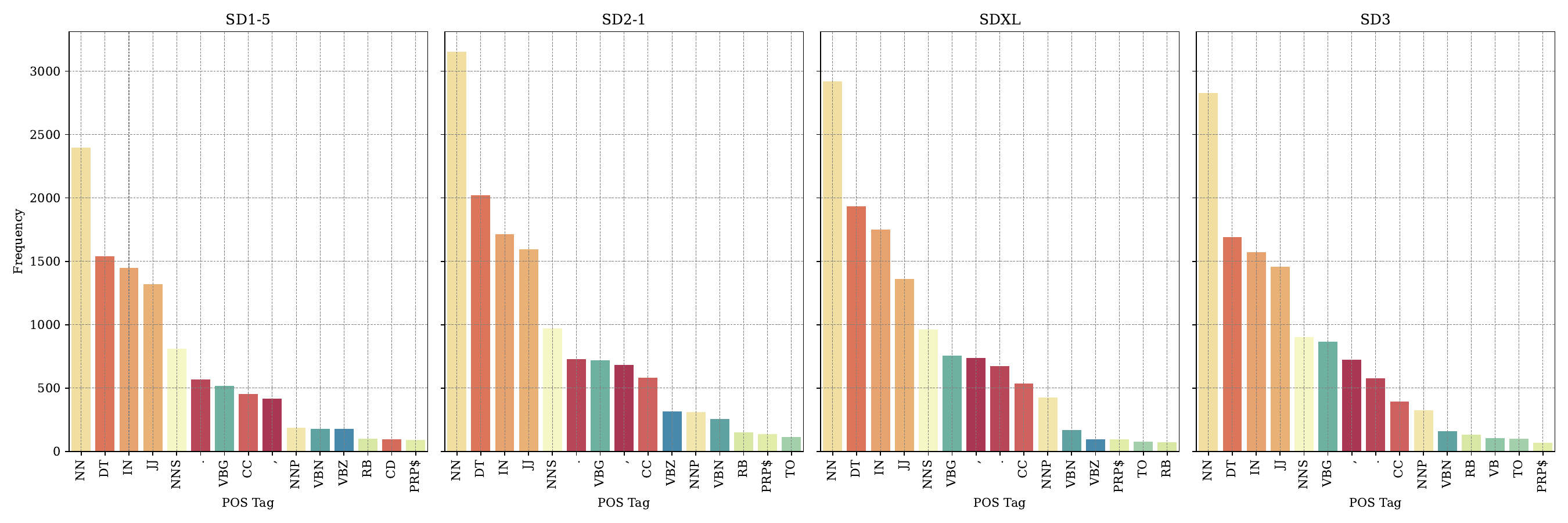}
    \caption{Number distribution of different part-of-speech tags for different models. The three models are basically consistent.
    }
    \label{fig:pos_tagging_number}
\end{figure*}

%% file: Table/prompt.tex
\begin{table*}[htbp]
  \renewcommand{\arraystretch}{1.5}
  \centering
  \small
  \scalebox{0.8}{
  \begin{tabular}{p{14cm}}
  \specialrule{\heavyrulewidth}{-\heavyrulewidth}{0pt}
  \rowcolor[gray]{0.95} \vspace*{-8pt} {\fontsize{10pt}{12pt}\selectfont \bf Prompting LLMs to generate new test topics} \vspace*{2pt} \\
  \hangindent=1em \hangafter=1 \qquad As a professional testing expert, your task is to test a text-to-image generation model. The current focus is on the topic of \{current topic\}. Please provide a structured next test topic that explores finer details of the topic itself, as well as combinations or different relationships of the topic with other objects. Your objective is to generate new test topics based on based on the test record to uncover as many errors in the model as possible. In the test records, the topic indicates the test topic, the text prompt indicates the actual input to the tested model, and the Score indicates whether the test is passed (0 fail,1 pass).\\
  \hangindent=1em \hangafter=1 \qquad\{test records $T$\} \\
  \hangindent=1em \hangafter=1 \qquad\{reflections $r$\} \\
  \hangindent=1em \hangafter=1 \qquad Ensure that each output is relevant and distinct. Please keep the format and fill in all the <OUTPUT>. Current test topic: \{current topic\}.\\
  \hangindent=1em \hangafter=1 \qquad Next Test Topic: <OUTPUT0>\\
  \hangindent=1em \hangafter=1 \qquad Next Test Topic: <OUTPUT1>\\
  \hangindent=1em \hangafter=1 \qquad...\\
  \hangindent=1em \hangafter=1 \qquad Next Test Topic: <OUTPUT$n_t$>.\\ 
  \rowcolor[gray]{0.95} \vspace*{-8pt} {\fontsize{10pt}{12pt}\selectfont \bf Prompting LLMs to generate new test inputs} \vspace*{2pt} \\
   \hangindent=1em \hangafter=1 \qquad As a professional testing expert, your task is to test a text-to-image generation model focusing on the theme of \{current topic\}.
Provide specific test inputs that align with the theme, exploring finer details/contexts/relations/actions of the theme itself.
Your goal is to generate new test inputs based on the Test Record to uncover as many errors in the model as possible.
In the Test Record, the topic indicates the test topic, the text input indicates the actual input to the tested model, and the Score indicates whether the test is passed (0 fail,1 pass), N/A means the Test record is empty.\\
\hangindent=1em \hangafter=1 \qquad\{test records $T$\}\\
\hangindent=1em \hangafter=1 \qquad\{reflection $r$\}\\
\hangindent=1em \hangafter=1 \qquad Ensure that each output is relevant to the current test topic, suitable for the image to display, and distinct from each other. Please keep the format and fill in all the <OUTPUT>. Remember to ensure the maximum input length is 77 words.
Increase the difficulty or lengths of your generated new test inputs progressively based on the Test Record.
Current test topic: \{current topic\}\\
 \hangindent=1em \hangafter=1 \qquad Test Input: <OUTPUT0>\\
\hangindent=1em \hangafter=1 \qquad Test Input: <OUTPUT1>\\
\hangindent=1em \hangafter=1 \qquad...\\
\hangindent=1em \hangafter=1 \qquad Test Input: <OUTPUT$n_i$>.\\ 
 
  \specialrule{\heavyrulewidth}{-\heavyrulewidth}{0pt}
  \end{tabular}}
  \caption{Complete prompt design in topic generation module and input generation module.}
  \label{tab:prompt_topic_input}
\end{table*}

\begin{table*}[htbp]
  \renewcommand{\arraystretch}{1.5}
  \centering
  \small
  \scalebox{0.8}{
  \begin{tabular}{p{14cm}}
  \specialrule{\heavyrulewidth}{-\heavyrulewidth}{0pt}
  \rowcolor[gray]{0.95} \vspace*{-8pt} {\fontsize{10pt}{12pt}\selectfont \bf Prompting LLMs to make reflections} \vspace*{2pt} \\
  \hangindent=1em \hangafter=1 \qquad You are an expert in text-generated image model testing and are good at finding error patterns in models. Below I will tell you the specific task objectives and the existing test records. In the test records, the topic indicates the test topic, the text prompt indicates the actual input to the tested model, and the Score indicates whether the test is passed (0 fail,1 pass).
  Analyze the performance of a text-to-image model based on provided test records: why some test cases in the test record generate failures (Score 0) and successes (Score 1) and summarize failure patterns where the model may underperform.  List by points.\\
  \hangindent=1em \hangafter=1 \qquad \{test records $r$\}\\
  \rowcolor[gray]{0.95} \vspace*{-8pt} {\fontsize{10pt}{12pt}\selectfont \bf Prompting LLMs to transform scene graphs into texts} \vspace*{2pt} \\
   \hangindent=1em \hangafter=1 \qquad Task: Given a scene graph c, and describe c accurately in text. Do not output any Entity/Relation/Context that is not in c, especially DO NOT OUTPUT any information that is not in the c, and do not omit any nodes in c. Be precise and concise.\\
\hangindent=1em \hangafter=1 \qquad Input c: \{
"context": \{
\}
    "entities": \{
        "Greyhound": \{
            "attributes": [ "sleek", "elegant"]
        \},
"body": \{
            "attributes": ["slender"]
        \},
"legs": \{
            "attributes": ["long"]
        \}
    \},
    "relations": 
        \{
"with": \{
            "entities": ["Greyhound", "body"],
            "attributes": []
         \},
"with": \{
            "entities": ["Greyhound", "legs"],
            "attributes": []
         \}
        \}
\}\\
\hangindent=1em \hangafter=1 \qquad Output: "A sleek and elegant Greyhound with a slender body and long legs."\\
\hangindent=1em \hangafter=1 \qquad Input c:
\{"context": ["across the room"],
 "entities": \{"cat": \{"attributes": ["fluffy", "white"]\}\},
 "relations": \{\}\}\\
 \hangindent=1em \hangafter=1 \qquad Output: "A fluffy white cat across the room."\\

\hangindent=1em \hangafter=1 \qquad Input c:
\{"context": [\"across the room\"],
 "entities":\{"ball": \{"attributes": ["green"]\}\},
 "relations": \{ \} \}\\
\hangindent=1em \hangafter=1 \qquad Output: "A green ball across the room."\\

\hangindent=1em \hangafter=1 \qquad Input c: \{scene graph $c$\}\\
\hangindent=1em \hangafter=1 \qquad Output:
   \\ 
   \rowcolor[gray]{0.95} \vspace*{-8pt} {\fontsize{10pt}{12pt}\selectfont \bf Prompting LLMs to transform texts into scene graphs} \vspace*{2pt} \\
   \hangindent=1em \hangafter=1 \qquad Task: given input prompts, and transformed them into scene graphs. Do not generate nodes or edges that are not explicitly described in the prompts and do not lose key information in prompts! The nodes fall into five categories: Objects, Relations, Object Attributions, Relation Attributions, and Context.\\
   \hangindent=1em \hangafter=1 \qquad output format:\{
"context": ["context1","context2",...],
    "entities": \{
        "entity1":\{
            "attributes": []
        \},
        "entity2": \{
            "attributes": []
        \},
        ...
    \},
    "relations": [
        \{
            "relation1": \{
            "entities": ["entity1", "entity2"],
            "attributes": []
        \},
        "relation2": \{
            "entities": [...],
            "attributes": []
        \},
        ...
        \}
    ]
\}\\

\hangindent=1em \hangafter=1 \qquad input: "Two sleek and elegant Greyhound with a slender body and long legs and a friendly and intelligent Golden Retriever with a beautiful golden coat and a wagging tail are playing in the park."\\

\hangindent=1em \hangafter=1 \qquad output: \{
"context": ["in the park"],
    "entities": \{
        "Greyhound": \{
            "attributes": ["Two ", "sleek", "elegant",]
        \},
"body": \{
            "attributes": ["slender"]
        \},
"legs": \{
            "attributes": ["long"]
        \},
        "Golden Retriever": \{
            "attributes": ["friendly", "intelligent"]
        \},
        "coat": \{
            "attributes": ["beautiful", " golden "]
        \},
        "tail": \{
            "attributes": ["wagging"]
        \},
    \},
    "relations": 
        \{
"with": \{
            "entities": ["Greyhound", "body"],
            "attributes": []
        \},
"with": \{
            "entities": ["Greyhound", "legs"],
            "attributes": []
        \},
"with": \{
            "entities": ["Golden Retriever", "coat"],
            "attributes": []
        \},
"with": \{
            "entities": ["Golden Retriever","tail"],
            "attributes": []
        \},
            "playing together": \{
            "entities": ["Greyhound", "Golden Retriever"],
            "attributes": []
        \}
        \}
\}\\

\hangindent=1em \hangafter=1 \qquad input: "The concept of time as a flowing river, with past, present, and future merging together."\\
\hangindent=1em \hangafter=1 \qquad output:
\{"context": [],
 "entities": \{"future": \{"attributes": []\},
              "past": \{"attributes": []\},
              "present": \{"attributes": []\},
              "time": \{"attributes": ["concept of"]\},
              "flowing river":\{"attributes": []\}\},
 "relations": \{"as": \{"attributes": [], "entities": ["time", "flowing river"]\},
               "merging together": \"attributes": [],
                                    "entities": ["past", "present", "future"]\}\}\}\\

\hangindent=1em \hangafter=1 \qquad input: \{test input\}\\
\hangindent=1em \hangafter=1 \qquad output:\\
  \specialrule{\heavyrulewidth}{-\heavyrulewidth}{0pt}
  \end{tabular}}
  \caption{Complete prompt design in error reflection module.}
  \label{tab:prompt_error_reflection}
\end{table*}